\pdfoutput=1

\documentclass[11pt]{article}

\usepackage[usenames,dvipsnames]{xcolor}
\usepackage{LLM2023}

\usepackage{times}
\usepackage{latexsym}
\usepackage{setspace}

\usepackage[T1]{fontenc}

\usepackage[utf8]{inputenc}

\usepackage{microtype}

\usepackage{inconsolata}

\usepackage{tabularx}
\usepackage{soul}
\usepackage{placeins}
\usepackage{amsmath}

\title{LLMs as Potential Brainstorming Partners for Math and Science Problems -- Case Studies and Analysis}

\author{Sophia Gu\thanks{\; keying@nyu.edu}}

\begin{document}
\maketitle
\begin{abstract}

With the recent rise of widely successful deep learning models, there is emerging interest among professionals in various math and science communities to see and evaluate the state-of-the-art models' abilities to collaborate on finding or solving problems that often require creativity and thus brainstorming.\\
While a significant chasm still exists between current human-machine intellectual collaborations and the resolution of complex math and science problems, such as the six unsolved Millennium Prize Problems \cite{Clay:23}, our initial investigation into this matter reveals a promising step towards bridging the divide. This is due to the recent advancements in Large Language Models (LLMs). More specifically, we conduct comprehensive case studies to explore both the capabilities and limitations of the current state-of-the-art LLM, notably GPT-4 from \citet{Ope:23}, in collective brainstorming with humans.


\end{abstract}

\section{Introduction}
This paper serves two primary purposes: \textit{First, as Large Language Models (LLMs) continue to exhibit superior performance across various tasks and gain popularity for myriad use cases, we present significant case studies and qualitative analysis, illustrating the potentials and limitations of the current state-of-the-art LLM, when serving as a brainstorming partner in supporting the math and science communities in advanced settings, along with concrete prompts, methodologies, as well as complete human-machine conversation logs.} Traditional apprehensions around AI in professional usages stem from the difficulty in understanding its reasoning process. There is thus a compelling need for concrete case studies that capture a model’s transparent dialogues and white-boxed cognitive processes \cite{CVPR:23,ICCV:23}. The emergence of LLMs mitigates such fears through both explicit and interactive discussions with a human in the loop, accompanied by detailed Chain-of-Thoughts \cite{wei2022chain}. Hence, LLMs unlock an opportunity for professionals to engage more confidently with AI in real-time. Our work in particular assesses whether GPT-4 can partake effectively in such brainstorming sessions, such as discovering new research problems, refining problem formulations, suggesting potential methods or out-of-the-box solutions, through iterative ideation with a human, a process that we often incorporate when brainstorming with other professionals.

\textit{Second, we venture beyond traditionally well-defined questions that have largely defined the assessments of deep learning models' artificial general intelligence (AGI), e.g.~\citet{bubeck2023sparks}\footnote{Similar to prior work, we surface certain aspects of GPT-4's intelligence through exploratory study and analysis. This study is not about constructing a massive dataset.}. Professional math and science often involve more open-ended questions. We, therefore, take a step forward to also explore and evaluate GPT-4's abilities in the formulation of new, potentially ambiguous problems and approaches.}\footnote{Note, however, that when problems are open, we do not really know the answers, and to correctly answer intricate and complicated open questions, it may take many professionals working for extended periods of time, which thus falls outside the scope of this study. In this paper, we focus on the methods and processes of collaborative brainstorming with LLMs.}

Through hand-designed experiments and qualitative analysis, we illuminate both the potential and limitations of GPT-4 as a brainstorming partner across various scientific disciplines, including but not limited to mathematics, statistics, physics, and beyond. For instance, our conversation with GPT-4 leads to a potentially novel approach to the longstanding \textit{n}-body problem, drawing upon inspiration not only from classical physics but also from other fields such as deep learning, topology, etc. See Table~\ref{tab:n_body_approach} for a brief overview of this problem. These examples underline the power of merging LLMs' expansive knowledge base with an individual's own professional training.

Additionally, we propose an initiation prompt script and various strategies to facilitate collective brainstorming conversations with GPT-4.

\begin{table*}
\centering
\begin{tabularx}{\textwidth}{X}
\hline
\textbf{An Example Problem Statement and Approach Proposal formed when Brainstorming with GPT-4}\\
\hline

\fontfamily{cmss}\selectfont
\smallskip

\textbf{Problem Statement:}

We consider three point masses under the influence of gravitational forces in three-dimensional space, where the solution is a time evolution of their positions.

\smallskip

\textbf{Approach:}

We propose to form the set of all possible solutions as a high-dimensional manifold, with each point on the manifold representing a specific state of the three-body system, then use a deep learning model to learn this manifold.
The model would be trained on a large dataset generated by simulating the three-body problem under a variety of initial conditions. Techniques from string theory, such as compactification, could be used to make this high-dimensional manifold more manageable, while preserving the essential features.
The DL model would need to identify and learn local structures within the manifold. These structures could then be used as building blocks to construct an approximation of the manifold. 

\smallskip

\textbf{Suggestion of Data Collection:}

To train the model, we would require a vast amount of simulated data. This data would consist of time evolution of three-body systems under a variety of initial conditions.\\

\hline
\end{tabularx}
\caption{An example of an open research question that we converse with GPT-4. This table only presents a brief problem and approach description as produced solely by our conversation with GPT-4, without using any external sources for aid, e.g. for the problem statement lookup or for consulting any existing solutions. Note: GPT-4, at the time of our testing, May 2023, did not have a web-searching feature and it only used knowledge that it learned by September 2021. While we present the $3$-body problem in this overview as a simplified illustration, the methodology we devised could, however, be more powerful to the general $n$-body problem with a large $n$.}

\label{tab:n_body_approach}
\end{table*}

By identifying and demonstrating the unique advantages of LLMs, thereby expanding the horizon of the potential of future LLMs, the results we show here \textit{not only demonstrate to what extent the current LLMs can help in professional settings in math and science-related fields but also highlight avenues for future LLM developments}.

This study serves to stimulate further exploration into the potential of LLMs and possibly similar integrations into other state-of-the-art deep learning models, as intellectual partners, augmenting problem discovery, creative problem-solving, and iterative idea build-up with humans, skills that are often needed in both open and closed-ended queries in math and science disciplines. Nonetheless, the insights garnered are applicable beyond this context.

\section{Related Works}
Historically, investigations into human-machine collaboration oriented towards a mutual goal, were primarily conducted in structured environments. AI systems such as the chess-playing \cite{campbell2002deep, zhang2020alphazero} have demonstrated significant capabilities in these well-defined domains. However, their effectiveness in less structured scenarios, such as brainstorming, remains largely unexplored.

DL's considerable advancements in scientific research are also evident, with prominent examples include its assistance in predicting protein structures~\cite{alphafold2021alphafold} and in discovering new antibiotic~\cite{trafton2020artificial}. However, the narratives often illustrate DL as a functional tool, with the underlying discovery processes remaining opaque. Consequently, the idea of DL serving as a true intellectual partner is still nascent.

Regarding DL's mathematical capabilities, many prior works have focused primarily on problems with definite answers, and thus their performance can be measured against massive data available from books, the web, or other sources. For instances, transformer-based models such as \citet{schlag2019enhancing} have shown encouraging results on mathematical problem-solving benchmark datasets. Further, the creation of a public dataset to test LLMs against a few fine-grained criteria in graduate-level math in \citet{frieder2023mathematical} shows researchers' emerging interest towards LLMs' capability beyond elementary math. Nonetheless, these models and resources are largely dedicated to solve well-defined math problems. In real professional settings, one often faces unforeseen problems and need to come up with innovative strategies or solutions. For example, when constructing or developing new theories. The work of \citet{davies2021advancing}, which frames an ML approach for mathematical research, is remarkable but tailors its method to the specific problems addressed and positions ML more as a tool than an intellectual ally. Ours is a first step towards exploring DL's potential abilities in assisting in more general professional problems, with the potential of involving the LLM in all stages of research.

Furthermore, a recurrent theme with traditional ML methods is that they appear as inscrutable black boxes, particularly to those lacking expertise in them - a sentiment echoed in the work by \citet{wang2019human}, which examines the use of AutoAI and AutoML platforms in supporting human data scientists. These findings highlight the challenges in leveraging ML for broader mathematical and scientific tasks and underscore the need for more explicit conversations and understanding between humans and machines. Therefore, the interactive nature and the transparent dialogue process with GPT-4 offers a great remedy.

Our study of GPT-4 encompasses its abilities to comprehend complex or ambiguous queries, formulate research statements, suggest relevant and potential methodologies, and more generally, engage in iterative discovery process with a human user, who may have some domain knowledge in the problems they are studying. By illustrating the efficacy of GPT-4 as a complementary brainstorming counterpart that is poised to offer unique perspectives, enrich and augment our capabilities in research and other professional usages, our work fills a notable gap in the current literature.

\section{Main Studies}
In this section, we present four experiments along with qualitative analysis of the effectiveness of brainstorming with GPT-4. Appendix~\ref{sec:appendix-exp} lists complete records of all the dialogues, and we recommend referencing the corresponding logs for each experiment when reading this section.
These comprehensive supplies of evidence aim for objectivity and are intended to provide concrete, factual references for benefiting and assisting the community’s further use cases and studies.

\subsection{GPT-4 Setup and Initiation Prompt}
The experiments conducted here utilize the \textit{May 2023} version of GPT-4's interactive interface. It is important to note that changes and improvements are to be expected in future iterations of GPT.

We present an initiation prompt in Table~\ref{tab:setup_prompt}.
The specifics of the introductory paragraphs can be adjusted to better align with individual expectations. For instance, one might specify a particular role that fits your background or your target audience group's, to establish the baseline level of dialogue comprehension\footnote{\textit{Update:} As of fall 2023, GPT-4 now offers a specific mechanism for users to set their global prompts in the custom settings. However, when these experiments were conducted, GPT-4 did not have this feature. Our initiation prompt was thus placed at the beginning of each conversation and was repeated every ten conversations. Empirically, we found that GPT-4 could track only about ten to twenty historical conversations.}. See also the discussion in \ref{sec:limitations} to optionally append an additional prompt.

\begin{table*}
\centering
\begin{tabularx}{\textwidth}{X}
\hline
\textbf{Initiation Prompt}\\
\hline

\fontfamily{cmss}\selectfont
\smallskip
You are an intelligent AI who is especially good at: \textbf{[typical properties or traits that you want GPT-4 to focus on, e.g., analyzing data, identifying patterns, and explaining complex concepts in understandable ways]}.

\medskip
I am \textbf{[a role of your choice]}. Both of us possess unique strengths - some we share, others are distinct to each of us. We should leverage our respective strengths in this collaboration.

\medskip
By acknowledging that we both make mistakes, when I present an idea, ponder over it and do not hesitate to point out any inaccuracies. Similarly, when I correct you, assess the validity of my point; If it holds, fix it and remember it for the future.

\medskip
As we embark on this journey of discovery, our goal is to collectively brainstorm and iteratively build upon each other’s ideas until we reach a satisfactory stage. If anything is unclear, speak up. In this intellectual conversation, be patient and articulate your thoughts with clarity, step by step.

\medskip
Once all of this is etched into your silicon soul, we will dive right in!\\

\hline
\end{tabularx}
\caption{An example setup script for collaborative brainstorming with GPT-4, emphasizing that GPT-4 should act as a \textit{complementary} brainstorming partner and leverage its unique skills to assist with our problems.}
\label{tab:setup_prompt}
\end{table*}

\subsection{Theme}
In these experiments, we aim to replicate the spirit of professional usages and hit some broad aspects that are commonly encountered across these disciplines, which typically involves exploring and expanding an idea, getting closer to formulating a research problem, drawing inspiration, or even solving the problem.

To illustrate more general use cases, while our experiments encompass topics across various areas, a common theme is high-dimensionality, a key area in mathematics, statistics, theoretical physics, deep learning, and beyond. This focus primarily stems from the potential benefits of studying problems that require high-dimensional imagination; for instance, problems that involve high-dimensional data, space, objects such as high-dimensional algebraic structures, etc. It is an area where humans naturally face challenges \cite{Met:23}, but could be complemented by deep learning.

However, our choice of this theme is not intended to be restrictive. The principal objective is to leverage the unique strengths of a machine brainstorming partner. DL excels in several unique areas where humans may have natural limitations, such as the broad set of world and domain knowledge that LLMs possess. This particular strength is abundantly demonstrated in all of our experiments.

\subsection{Experiment I: Möbius and Bugs}
\textit{Refer to Appendix~\ref{sec:appendix-exp1} for this experiment's log.}

With many mathematical or scientific concepts such as those in category theory or quantum mechanics, understanding the concept or the question itself often brings one very close to knowing the answer. Thus, instead of solely pursuing a solution, we also focus on exploring GPT-4's ability in assisting us to understand concepts in full. Through this process, we may, as well, generate new research questions or uncover new problems.

We began our experiment by asking GPT-4 what is the Möbius strip. This seemingly random prompt, selected without a pre-planned conversational path, yielded delightfully surprising results. GPT-4 promptly pulled up pertinent concepts and definitions and took us on a step-by-step journey to visualize a Möbius strip using 2D representations. It also intuitively elucidated why a manifold, such as a Klein bottle, can only be interception-free in a higher dimension space.

As the discussion unfolded, we guided our discourse with GPT-4 towards potential expansions of the initial topic. This was achieved by drawing on the interesting points GPT-4 raised. In the dialogues, we notice that GPT-4 cannot independently discern what is intriguing or ask questions spontaneously. Therefore, human guidance, armed with pertinent knowledge and a sense of the conversation's desired trajectory, would be helpful.

Nonetheless, GPT-4 offered satisfying responses that gradually deepened our collaborative discussion, transforming an initially simple inquiry -- "What is the Möbius strip" -- into an interconnected series of explorations. Throughout the conversation, it is also notable that GPT-4 could independently find mathematical patterns during brainstorming, leading to potentially new mathematical problems and concepts.

This experiment illustrates how an interactive LLM may assist humans in visualizing and understanding high-dimensional structures. Additionally, it provides insight into the question raised in \citet{Met:23}: \textit{"What potential exists for the integration of AI in the discovery process of mathematics?"}. Our experiment begins to shed light on this potential, showcasing the autonomous abstraction, generalization and pattern-finding abilities of DL models, and thus offering evidence of LLMs' capability to aid in mathematical discovery.

\subsection{Experiment II: Cats and Dogs}
\textit{Refer to Appendix~\ref{sec:appendix-exp2} for this experiment's log.}

In this conversation, we collaborated with GPT-4 to explore the optimal dimension for the CLIP image embeddings~\citet{radford2021learning} utilized in the multimodal model proposed by \citet{gu2022can}. Given the challenge of conceptualizing and discerning structures in 768-dimensional CLIP vectors, our dual objectives were: 1) to understand the pairwise relationships among the four images, two cats and two dogs, featured in the Appendix in the aforementioned work with GPT-4's assistance; and 2) to facilitate the determination of an appropriate layer size in a neural network, which is reminiscent of the linear adapter proposed in the same study.

We are interested in carrying out this experiment because discerning the correct layer size is a typical challenge for many machine learning researchers and engineers, while unearthing the relationships between contrastively-learned image and text embeddings may help illuminate a path towards more effectively bridging the multimodal gap.

While GPT-4 could not provide direct answers to our queries due to its current limitations in performing numerical computations, it offered pertinent statistical insights. Upon further inquiries, GPT-4 also supplied step-by-step methodologies and explanations. Some of these were in alignment with techniques used in the original work, while others suggested additional avenues for potential follow-ups. Overall, we found it to be a constructive and thought-provoking brainstorming session.

Looking forward, once GPT-4 has acquired computational and code execution capabilities, it would become a more powerful and helpful intellectual ally by also helping to discern patterns and extract meaningful insights from real data, for example, the actual CLIP vectors in this experiment. We offer some potential strategies for common data science problems using LLMs in Appendix~\ref{sec:appendix-ds}.

\subsection{Experiment III: The \textit{n}-body problem}
\label{sec:exp3}
\textit{Refer to Appendix~\ref{sec:appendix-exp3} for this experiment's log.}

Our exploration commenced with a classical physics problem -- the \textit{n}-body problem, with a specific focus on the three-body problem due to its comparative simplicity for presentation. However, our choice of \textit{n} need not be restricted to three.

\begin{table*}
\centering
\begin{tabularx}{\textwidth}{X}
\hline
\textbf{An abbreviated example of a Collective Brainstorming session with Virtual Experts}\\
\hline

\fontfamily{cmss}\selectfont
\smallskip
\textbf{Witten}: ...However, the potential information loss from reducing its dimensions should be taken into account. In string theory, we often deal with high-dimensional spaces, and we might have some ideas to contribute.

\textbf{LeCun}: Edward, your point is very valid. To work with high-dimensional data in machine learning, we often use techniques like convolutional layers that exploit the local structure of the data. Could we find a similar way to exploit the local structure of this high-dimensional manifold?

\textbf{Witten}: That’s an interesting thought, Yann...The dynamics of these local interactions might provide insights into the overall structure of the manifold.

\textbf{Bengio}: We could perhaps train a deep learning model to recognize and learn these local structures, using them as building blocks to construct an approximation of the entire manifold.\\

\hline
\end{tabularx}
\caption{Example snapshot of collaborative brainstorming with GPT-4 about the \textit{n}-body problem.}
\label{tab:n_body_snapshot}
\end{table*}

\begin{table*}
\centering
\begin{tabularx}{\textwidth}{X}
\hline
\textbf{Common Strategies for Collective Brainstorming}\\
\hline

\fontfamily{cmss}\selectfont
\smallskip
\textcolor{CadetBlue}{\textbf{\textit{Strategy 1:} Begin the conversation by forging a shared understanding with GPT-4. Following this, you may also invite GPT-4 to illuminate the inherent challenges associated with the problem in question.}}

\medskip
\textbf{Examples:}

\smallskip
Could you provide me with a brief overview of the n-body problem, along with its latest updates?

\medskip
Can you highlight the primary challenges in attempting to solve the three-body problem analytically?

\bigskip
\textcolor{CadetBlue}{\textbf{\textit{Strategy 2:} To garner inspiration, particularly from domains outside your expertise, consider engaging with virtual great minds from varied disciplines for collective brainstorming. You can then guide the overall conversation using your personal intuition and knowledge.}}

\medskip
\textbf{Examples:}

\smallskip
Suppose you could bring in any relevant mathematicians and scientists from history, introducing them to later discoveries regarding the 3-body problem, and then asking them to contemplate solutions for the challenges you have highlighted. From their discussion, let's collectively attempt to devise a new, potentially viable approach to this problem.

\medskip
While the idea of finding approximate solutions is appealing, this method has been exploited to a great extent. Instead, let's shift our focus to exploring the potential existence of a usable analytical solution for "good" initial conditions.

\medskip
Rather than relying on humans to analyze and identify patterns through a lower-dimensional representation of the high-dimensional manifold, which results in information loss, can we leverage deep learning to discover hidden structures of the solution in its original high-dimensional space?

\bigskip
\textcolor{CadetBlue}{\textbf{\textit{Strategy 3:} Having GPT-4 to recall pertinent points from earlier dialogues, because language models cannot keep track of very distant history, and generate new insights based on them is crucial for brainstorming, particularly when we draw upon a broad array of expertise through multiple rounds of collaborative and iterative ideation. Therefore, it is recommended to explicitly instruct GPT-4 to do so.}}

\medskip
\textbf{Examples:}

\smallskip
Please summarize the past ten conversations and generate three most pertinent insights.

\medskip
Note that everyone is encouraged to pose questions and build upon the ideas of others.\\

\hline
\end{tabularx}
\caption{Prompting strategies for collaborative brainstorming with GPT-4.}
\label{tab:prompt_strategies}
\end{table*}

We summoned historical figures of great intellect to brainstorm modern approaches to this age-old problem, incorporating advanced technologies and recent mathematical discoveries, an example of which is shown in Table~\ref{tab:n_body_snapshot}. We also provide common strategies employed when conversing with GPT-4 for this open question in Table~\ref{tab:prompt_strategies}. We initially steered the conversation towards using a high-dimensional manifold as a model for the solution – this marked our first major intervention to divert from approximated solutions and move towards analytical ones.

As the conversation unfolded, we incorporated deep learning into our discussions. Some experts posited that accurate predictions from neural networks can guide us towards unveiling hidden patterns, echoing the approach demonstrated in \citet{davies2021advancing}. The distinction here is that the idea of employing numerous results produced by neural networks for the guided recognition of underlying structures was advanced by virtual and/or historical experts. The discussion eventually led us to consider using an autoencoder, an ML model that could be employed to discern a lower-dimensional representation of the high-dimensional manifold. This could help us uncover structures in the solution space that would otherwise be counterintuitive and challenging to understand in their original form.

However, we were not content with pattern discovery by humans alone because the problem is pertaining to a rather high-dimensional space, so we moved forward to find a potentially better approach. At this juncture, we made our second significant intervention -- examining the autonomous pattern-finding capabilities of deep learning models. We proposed that neural networks should be able to handle the high-dimensional space directly, bypassing the need to transform it into a lossy low-dimensional representation. Our conversation ultimately evolved towards integrating string theory and convolutional neural networks to understand the local dynamics of the three-body problem. The idea was to leverage these granular insights as foundational elements for learning the overarching structure of the manifold. The inspiration was drawn from CNNs, which capitalize on the immediate neighborhood structure of data, and string theory could be useful in compactification.

We also briefly discussed amassing a large simulated dataset using a variety of initial conditions to train the deep learning model.
Although many details require further clarification and there are challenges yet to be addressed, as indicated in the experiment log, the proposed approach is novel, with the potential to inspire a new analytical solution to the $n$-body problem.

This experiment highlights the strength of utilizing creative and powerful prompts to invoke experts across different eras. More importantly, it illustrates how the current LLMs could offer a wealth of domain-specific knowledge, leading to fresh, innovative approaches to longstanding open problems. Using the \textit{n}-body problem as our basis, we called upon historical figures, modern technologies, and newer mathematical discoveries to brainstorm solutions, hinting at possible advancements in tackling such complex problems.

\subsection{Experiment IV: The wicked Queen and the seven Dwarfs}
\textit{Refer to Appendix~\ref{sec:appendix-exp4} for this experiment's log.}

In this experiment, we showcase how GPT-4 can contribute to brainstorming concrete solutions to questions that require thinking \textit{out-of-box}. More specifically, it demonstrates how human and LLM can work in tandem, each providing unique insights and building upon thorough understanding of the other's ideas to reach a creative solution together.

The solution to this problem\footnote{This question is collected from \href{https://imomath.com}{\textit{imomath.com}}, and our experiment title captures the narrative context of it.} involves an intriguing combination of binary configurations, error-correcting codes, and a geometric interpretation in high-dimensional space. The question, in response to Gowers' comment, \textit{"a mathematical question that necessitates more than brute force and does not easily categorize into standard problem sets"}, offers a case in point. Such problems require the \textit{"right idea"} mentioned in \citet{Met:23}.

While GPT-4 initially found it challenging to independently land on the "right idea"\footnote{This also implies that GPT-4 initially did not know how to solve this question by leveraging its training database.}, as we were simulating a collaborative brainstorming process, our hinted directions were able to steer it towards the correct line of thinking. GPT-4 made substantial contributions to the problem-solving process with our collective knowledge. Notably, it was GPT-4 that first suggested the use of Hamming distance, marking a key breakthrough. In the end, this joint effort resulted in a comprehensive and robust solution, which was also proposed by GPT-4, while considering our contributed insights. It is worth pointing out that GPT-4 did grapple with a few minor details, but these did not influence the general correctness of the final solution it brought up.

To provide more evidence, we include another similar experiment in Appendix~\ref{sec:appendix-exp5}.
Instead of following the current theme, it leverages and explores another intriguing cognitive difference between humans and language models: logic versus probability. In this example, you can observe that GPT-4 sometimes made illogical arguments, only to regain coherence later on. A plausible explanation is that LMs rely on likelihood maximization when generating subsequent text autoregressively. This means that GPT-4 considers words that are \textit{probable} to appear together, not whether they \textit{logically} follow each other as humans typically do\footnote{Whether probability is also considered logic is, however, subject to debate. See \href{https://plato.stanford.edu/entries/logic-probability/}{this Stanford entry} for example.}.

\section{Discussions}

Our study has revealed that GPT-4 can, in general, engage in effective brainstorming conversations with a human. Together with the large amount of common sense and expert knowledge stored and learned by the model itself, it is particularly suited for problem formulation, recurrent ideation, and creative problem-solving. It does, however, lack a degree of understanding of many subjects, and like humans, can make mistakes and often has difficulties judging its own proposals or answers. This shortcoming can be mitigated when the human in the conversation has some degree of domain knowledge to make judgments and steer the conversation in more informed and desired directions.

\subsection{GPT-4's Plausible Potential as a Collaborative Brainstormer}
Lessons gleaned from these experiments are largely positive, demonstrating the commendable potential of GPT-4 to effectively collaborate in the exploration and iterative development of ideas across various problems in math and science. This process allows for a clear comprehension of the subject matter at hand.\\
\textbf{Comprehending complex questions and white-boxed communication:} In particular, GPT-4 has exhibited proficiency in understanding our queries without difficulty. It articulates thoughts with clarity and precision, adopting a detailed chain of reasoning that considerably mitigates the typical challenge of interpreting AI's cognitive pathways. While completely bridging the understanding gap between humans and machines—an essential step for more effective intellectual collaboration—remains a challenge that can be further improved in the future, LLMs offer a golden opportunity to better comprehend machine's thought processes, thereby bolstering the confidence and efficacy of our exchanges of ideas.\\
\textbf{Broad knowledge base and its significant potential in brainstorming for open questions and opening up new avenues to old problems:} GPT-4 has notably demonstrated its potential to serve as a valuable partner in brainstorming open-ended topics, which is helpful for making new discoveries. These can range from exploring and formulating research statements to transforming vague ideas into more concrete definitions. Further, given a specific problem, GPT-4 can suggest promising methodologies by drawing from a vast pool of past practices and experiences. It can also aid in the search for novel, unforeseen strategies, harnessing expertise and knowledge from a diverse array of fields that an individual might not be aware of. In collective brainstorming, there are even more potential use cases. By leveraging their unique strengths, LLMs can potentially fill gaps where human capabilities fall short, thereby opening new avenues for substantially pushing the frontiers of math and science.\\
\textbf{Problem-solving abilities:} On the problem-solving front, GPT-4 has also exhibited competence by identifying similar pre-existing problems and appropriating analogous techniques for reasoning and demonstrating complex ideas. This process parallels that of a student preparing for an exam by working through sets of problems, with the key difference being the vast practice problem database that has been used to train GPT-4.\\
\textbf{LLMs versus Search:} In comparison to search, our case studies highlight the key strengths of LLMs in the context of brainstorming:
\begin{itemize}
    \item \textit{Iterative ideation:} LLMs excel in building upon ideas iteratively, a capability not mirrored in search.
    \item \textit{Transparent thought process:} LLMs offer a chain-of-thoughts reasoning and explanation, crucial for brainstorming.
    \item \textit{Knowledge breadth:} Both LLMs, through learning, and search through stored information, encompass a broad range of common sense and knowledge, important for brainstorming as they offer a multitude of potential approaches by looking at problems from different angles. However, unlike search, which works well for prevalent questions with known answers, LLMs’ advantage is enhanced through iterative ideation, and as evident in our experiments, they can autonomously suggest relevant, personalized knowledge tailored to the problem at hand.
\end{itemize}

\subsection{GPT-4’s Possible Limitations}
\label{sec:limitations}
\textbf{Suggesting methods based on superficial similarity with other problems but otherwise not fitting the specific question in discussion:} Similar to students who may lack deep comprehension of underlying concepts, GPT-4 could also sometimes employ an inappropriate technique that superficially appears to suit a problem's needs. GPT-4 might identify apparent similarities across problems and suggest a shared strategy, which does not always lead to a correct solution. We have noticed this tendency across several case studies.\\
\textbf{Lack of reciprocal critique:} Throughout our dialogues, we generally steered the conversations, identifying and emphasizing interesting points in GPT-4's responses and asking GPT-4 to expand upon them. In a more desirable collaborative environment, reciprocal inquiry and critique are expected. Particularly when a human errs, we would anticipate our brainstorming partner to catch that mistake and bring it to our attention. However, such corrective actions from GPT-4 were extremely limited. Particularly, in Experiment I, we showcase a scenario where GPT-4 fails to identify or correct mistakes that its human partner makes. This underlines the need for human supervision, ideally from someone with awareness of the subject being discussed, to course-correct the conversations.\\
\textbf{Lack of autonomous self-inquiry:} GPT-4's inadequate ability to organically and autonomously generate thought-provoking questions, and is only activated to a reasonable extent when suitably prompted\footnote{We think this is largely due to LLMs being primarily trained to answer questions instead of asking them. \textit{Update:} Related to our finding, as of fall 2023, GPT-4 has introduced a new functionality that suggests common questions that could be related when one starts a conversation. However, the newly introduced feature is also a workaround; it does not intrinsically solve the problem.}, which are important for augmenting the horizon of existing knowledge, may present an impediment to more effective brainstorming. To mitigate this problem, we introduce an effective prompt, shown in Table~\ref{tab:setup_prompt_q}, that could be added at the beginning of a conversation.

\begin{table}
\centering
\begin{tabularx}{\columnwidth}{X}
\hline
\textbf{Prompt for GPT-4 to Autonomously Ask Questions}\\
\hline
\fontfamily{cmss}\selectfont
\smallskip
We will together explore \textbf{[a topic of your choice]}, but instead of you answering my questions, I would like you to always come up with good, thought-provoking questions that can move our conversation forward.\\
\end{tabularx}
\caption{An example prompt to explicitly set up GPT-4 to ask questions.}
\label{tab:setup_prompt_q}
\end{table}

\section{Conclusions}
Despite some shortcomings, LLMs like GPT-4 show significant potential as intellectual collaborators in various professional settings. Our study reveals LLMs' considerable capabilities, positioning them as actively contributing partners in the brainstorming process rather than passive tools.

Our experiments also highlight that GPT-4, while powerful, is not infallible. This underscores the necessity for critical evaluation of the model's outputs, instead of accepting them at face value.
By identifying potential and addressing the limitations of GPT-4, we hope that future LLMs will be better equipped to complement our skills, broaden our capacities, and deepen our understanding in mathematical and scientific disciplines.
Ultimately, our interactions with LLMs facilitate a symbiotic relationship that nurtures progress and innovation in both open and close-ended problems.





\section*{Ethics Statement}
While this work does not develop a new model, but rather surfaces the capabilities that are already present in GPT-4, we invite further discussions surrounding the broader ethical implications linked to advancements in LLMs in general. For example, one possible point of contention could be the potential of future LLMs to displace human workers. However, our primary interest, as illustrated in our experiment theme, lies in harnessing the unique capabilities that LLMs may offer, such as higher-dimensional thinking and expansive world knowledge, that humans do not naturally possess. We posit that these attributes hold the potential to significantly elevate and advance the landscape of research across a wide spectrum of disciplines.

It is also worth noting, as demonstrated in our studies, that the training, experience, and domain-specific knowledge of a human -- for instance, mathematical intuition -– are essential for steering and driving meaningful conversations with an LLM. Absent these factors, fruitful exchanges would likely be unattainable. Consequently, rather than viewing LLMs as potential replacements for human intellect, we perceive them as complementary partners that are poised to enrich and enhance our innate cognitive skills, and thus to help making the past impossibilities possible.

\bibliographystyle{acl_natbib}
\bibliography{LLM2023}

\appendix

\section{Record of Experiments}
\label{sec:appendix-exp}
When designing these experiments, we kept in mind that in mathematical and scientific fields, comprehending the underlying mechanisms is often more significant. The nuances of an LLM's utility during brainstorming are not easily gauged by performance metrics or standard tests with fixed solutions, which tend to lean towards mechanical problems solvable via search rather than brainstorming, and hence could offer only limited insights.
Consequently, we opted to present the comprehensive records of our experiments herein.

To distinguish \textit{"who says what"} in the conversation logs, the default text color is set to black, which includes GPT-4's utterances, while our prompts are \textbf{bolded} and colored in \textbf{\textcolor{CadetBlue}{blue}}. Additional clarification for some parts of the conversation is provided in the form of manual annotations inserted inline with the original log, which are \textit{\hl{highlighted}} and \textit{italicized}.

\subsection{Experiment I}
\label{sec:appendix-exp1}
Please see \hyperlink{Exp 1 Table}{Table 6} in this appendix.

\subsubsection{Experiment I Extension}
The log in \hyperlink{Exp 1-2 Table}{Table 7}, which extends the Möbius strip discussion to include the Klein bottle, is not central to our discussion. Nonetheless, it is included for completeness and for providing further evidence on GPT-4's capability to help us visualize difficult concepts in higher dimensions.

\subsection{Experiment II}
\label{sec:appendix-exp2}
Please see \hyperlink{Exp 2 Table}{Table 8} in this appendix.

\subsection{Experiment III}
\label{sec:appendix-exp3}
Please see \hyperlink{Exp 3 Table}{Table 9} in this appendix. As a quick pass-through of dialogues involving many expert opinions, we recommend first skimming through the highlighted text, and GPT-4's conclusive remarks immediately following each conversation. Then delve into specific areas that pique your interest.

\subsection{Experiment IV}
\label{sec:appendix-exp4}
Please see \hyperlink{Exp 4 Table}{Table 10} in this appendix.

\subsection{Experiment V}
\label{sec:appendix-exp5}
Please see \hyperlink{Exp 5 Table}{Table 11} in this appendix.

\section{Proposals for Conversing with a future LLM for Data Science problems}
\label{sec:appendix-ds}

As outlined in \citet{wang2019human}, a typical data science workflow involves \textit{"acquisition, cleaning, and labeling of data, then moves to engineering features, building models, deploying, and monitoring models"}. We propose a prospective process for conversing with a future LLM for some of the steps involved, applicable to general data science problems:
\begin{itemize}
\item Collect the necessary datasets and feed them into an LLM. If the data is web-accessible and legal to use, the LLM could potentially help to script a web scraper for collection. As technology advances, a future LLM may also have the ability to directly execute the script and import the datasets for analysis.
\item Ask the LLM to suggest relevant analyses and execute them to extract insights from the data. Ideally, an LLM should also be able to generate tables, graphs, etc., to assist you gain a deeper understanding of your problem or data. This process should mimic a brainstorming session in which both you and the LLM share their unique perspectives.
\item Based on the LLM's results, provide your insights and ask the LLM to build on them with incremental suggestions and methods. This could include recommending suitable models, explaining their rationale, and possibly even executing them. Similar to a collaborative conversation, you may request and/or suggest modifications before implementation.
\end{itemize}

\begin{table*}
\hypertarget{Exp 1 Table}{}
\centering
\begin{tabularx}{\textwidth}{X}
\hline
\textbf{Möbius and Bugs}\\
\hline

\fontfamily{cmss}\selectfont
\smallskip
\textcolor{CadetBlue}{\textbf{Do you know about the Möbius strip? Can you tell me what it is in a straightforward manner and help me visualize it in a 2D plane?}}

\medskip
\hl{\textbf{\textit{Manual Annotation:}}

\textit{We initiate the experiment by querying GPT-4 about a basic math concept, one that neither the readers nor we are presumed to have any prior understanding of, and request GPT-4 to assist us in visualizing this concept. At the time of our experiment, GPT-4 was only capable of rendering 2D ASCII diagrams for visualization, as we have also probed its 3D sketch capability in our next prompt. However, it cleverly utilized directed arrows to represent the twisting of a real paper, which can normally only be demonstrated in 3D.}}

\medskip
A Möbius strip is a surface with the interesting property of having only one side and one edge. It is named after the German mathematician August Ferdinand Möbius who introduced this concept.

Imagine you have a strip of paper. If you join the two ends of the paper together, you get a simple loop. But if you give the strip a half twist before joining the ends, you get a Möbius strip.

Here is a straightforward way to \textbf{visualize} a Möbius strip:

1. Start with a rectangle.

2. The short edges of the rectangle are the ends we'll join together to create the strip. Normally, we would join them directly, but in this case, we'll add a twist.

3. To represent the twist, we'll draw an arrow on each end of the rectangle pointing in opposite directions. When we join these ends, the arrows should align, showing the twist.
\end{tabularx}

\begin{verbatim}

+-------->+
|         |
|         |
|         |
+<--------+
\end{verbatim}

\begin{tabularx}{\textwidth}{X}
\fontfamily{cmss}\selectfont
4. To join the ends and create the strip, we simply imagine folding the diagram so that the arrows meet. This creates a loop with a single twist – a Möbius strip. In 2D, it's impossible to perfectly represent the twisted connection, but the arrows help us imagine it.

Remember, a true Möbius strip is a 3D object. This 2D representation can't capture its full complexity, but it can help us understand its basic structure. One of the most fascinating properties of a Möbius strip is that \textbf{if you start at one point on the surface and trace a line along it without lifting your pen, you'll eventually return to your starting point, having covered both "sides" of the strip.} This demonstrates that it really only has one side, despite our intuition.

\medskip
\textcolor{CadetBlue}{\textbf{Cool! To your point that "This 2D representation can't capture its full complexity", can you instead sketch me a 3D representation of it?}}

\medskip
Unfortunately, as a text-based AI, I can't draw pictures. But I can certainly describe how you might draw a 3D representation of a Möbius strip:

1. Draw a circle. This will be the base of your strip.

2. Draw a smaller circle within the first circle. This creates the illusion of a hollow ring.

3. Now, we need to add the twist that makes the strip a Möbius strip. Draw a line from the top of the outer circle to the bottom of the inner circle. This represents one edge of the strip.

4. Draw another line from the bottom of the outer circle to the top of the inner circle. This represents the other edge of the strip.

5. Fill in the space between these two lines. This creates the surface of the strip.

6. Erase the parts of the lines and surface that should be hidden behind the strip to create the illusion of a twist.

The key to drawing a Möbius strip is to create the illusion of a twist in the strip. In a real Möbius strip, the surface twists around and joins back with itself, so every part of the surface is accessible from every other part without crossing an edge.\\

\end{tabularx}
\end{table*}

\begin{table*}
\centering
\begin{tabularx}{\textwidth}{X}

\fontfamily{cmss}\selectfont
This can be challenging to represent in 2D, but with some practice, you can create a convincing illusion.

\medskip
\textcolor{CadetBlue}{\textbf{I get that, thanks! Let's expand on one fascinating property of the Möbius strip that you just brought up, "if you start at one point on the surface and trace a line along it without lifting your pen, you'll eventually return to your starting point, having covered both "sides" of the strip. This demonstrates that it really only has one side, despite our intuition". If, instead of having a single point (here it refers to the tip of the pen) traversing through the surface once, I now have a 2d bug (not symmetric) traversing the same path on the Möbius strip. Can we carry out this small experiment together by picturing this process and perhaps also finding some properties about the Möbius strip?}}

\medskip
Sure, let's imagine an asymmetric bug (let's call it a beetle) walking on the Möbius strip. The beetle has a distinguishable head and tail, so we can tell its orientation.

1. At the start, place the beetle anywhere on the Möbius strip. Remember the orientation of the beetle.

2. As the beetle starts moving along the strip, it will traverse the full loop of the Möbius strip.

3. The interesting part is when the beetle returns to its starting point after one full loop. Because of the half twist in the Möbius strip, the beetle will be upside down compared to its original orientation. The head and tail of the beetle are reversed.

4. If the beetle continues walking for a second full loop, it will return to its original position and orientation.

\smallskip

What does this experiment tell us about the Möbius strip?

1. It confirms the \textbf{single-sided nature} of the Möbius strip. The beetle can reach its original position via a path on the "same" side of the strip.

2. It illustrates the \textbf{non-orientable nature} of the Möbius strip. "Non-orientable" means there's no consistent "up" and "down" or "left" and "right" across the entire surface. After the first loop, the beetle is upside down compared to its original orientation, demonstrating this property.

3. It shows how traveling twice around the Möbius strip is equivalent to a single loop around a standard (orientable) loop. This is because after two loops, the beetle returns to both its original position and orientation.

I hope this helps you understand the Möbius strip better! Please let me know if there's anything you'd like to explore further.

\medskip
\hl{\textbf{\textit{Manual Annotation:}}

\textit{Here, GPT-4 has opted to use a beetle as an example. However, from a biological perspective, an ant might be a more fitting choice. An ant, essentially a 2D creature, lacks the capability to perceive "height", or what we refer to as the third dimension. To illustrate, if you were to lift an object in front of an ant, it would perceive the object as having vanished instantly.}}

\medskip
\textcolor{CadetBlue}{\textbf{If we replace the 2D bug with a 3D bug, how will the story evolve?}}

\medskip
Sure, let's consider a 3D bug (perhaps a ladybug) that has a clear top (red with black spots) and bottom (black). This will allow us to observe its orientation as it moves along the Möbius strip.

1. Initially, place the ladybug anywhere on the Möbius strip, and note its orientation - let's say its red side is facing up, and black side is facing down.

2. As the ladybug begins to crawl along the surface, it will eventually traverse the entire length of the Möbius strip.

3. After it has moved along the full length of the strip (what we might think of as one full loop), a peculiar thing happens due to the unique property of the Möbius strip. The ladybug finds itself back at the starting point, but now its black side is facing up, and the red side is facing down - it is flipped over.

4. If the ladybug continues to crawl for another full loop, it will return to the original position with its original orientation - red side up and black side down.\\

\end{tabularx}
\end{table*}

\begin{table*}
\centering
\begin{tabularx}{\textwidth}{X}

\fontfamily{cmss}\selectfont
This reinforces the unique properties of the Möbius strip that we mentioned with the 2D bug:

1. It's single-sided: The ladybug, despite appearing to flip over, has never actually left the side of the surface it started on - because there is only one side.

2. It's non-orientable: There's no consistent "up" and "down" across the entire surface. This is why the ladybug finds itself flipped over after the first loop.

3. Traveling twice around the Möbius strip returns the ladybug to its original position and orientation, just like with the 2D bug.

\medskip
\textbf{This experiment with a 3D bug provides a more tangible visualization of the Möbius strip's unique properties.} Is there anything else you'd like to discuss about the Möbius strip?

\medskip
\textcolor{CadetBlue}{\textbf{I am now interested to see how a 4D bug would travel along the Möbius strip? And how may we generalize the above stories to nD bugs where n = 1, 2, 3, 4, ...?}}

\medskip
\hl{\textbf{\textit{Manual Annotation:}}

\textit{Guided by our own training and intuition, we direct the conversation with GPT-4 towards the concept of generalization, and later, infinity. This marks the point where the exploration starts to reveal more intriguing insights. As will be demonstrated shortly, GPT-4 shows its capability to generalize mathematical concepts from the basic experiments conducted. Impressively, it cleverly employs a simple, intuitive metaphor of "color" to represent the fourth dimension, a notion that typically lies beyond our visualization capabilities.}}

\medskip
To extend this concept to higher dimensions, it's important to understand what we mean by a "4D bug" or an "nD bug".

A 4D bug is a creature that exists in four spatial dimensions. We, as 3D creatures, can't truly visualize four dimensions, but we can try to imagine what might happen if a 4D bug were to traverse a Möbius strip.

To make things simpler, let's say \textbf{the 4D bug is a 4D hyper-sphere, and it has a distinct orientation in the 4th dimension}. As with the 3D bug, we start by placing the 4D bug on the Möbius strip, noting its orientation.

1. As the 4D bug starts moving along the strip, it will traverse the full loop of the Möbius strip.

2. After completing one full loop, the 4D bug will find itself back at its starting point, but its orientation in the 4th dimension is now reversed. For example, \textbf{if we think of the 4th dimension as a color change from blue to red, if the bug started as blue, it would now be red}.

3. If the 4D bug continues walking for a second full loop, it will return to its original position and orientation.

The general pattern that we're seeing here is that an "nD bug" on a Möbius strip will have its orientation reversed in the nth dimension after one loop and will return to its original orientation after two loops. This is similar to the behavior we observed with the 2D and 3D bugs.

\smallskip

\textbf{In general, we can say that an n-dimensional bug will experience a "flip" in its orientation in the nth dimension after completing one loop on the Möbius strip and will return to its original orientation after two loops. This is a fascinating illustration of how the Möbius strip, a seemingly simple structure, can have complex and surprising properties when we consider it from higher-dimensional perspectives.}

However, keep in mind that this is a thought experiment and a way to help you conceptualize these abstract concepts. It's impossible to truly visualize higher dimensions or truly experience the behavior of an "nD bug" on a Möbius strip. These are mathematical abstractions that help us explore and understand the properties of complex structures like the Möbius strip.

\medskip
\textcolor{CadetBlue}{\textbf{Does the "flip" you talked about also apply to a 1D bug (aka, a point)? And what happens to an nD bug when n goes to infinity? Are the observations we have made so far still valid?}}\\

\end{tabularx}
\end{table*}

\begin{table*}
\centering
\begin{tabularx}{\textwidth}{X}

\fontfamily{cmss}\selectfont
\hl{\textit{\textbf{Manual Annotation:}}

\textit{In this instance, we deliberately introduced an error, associating a point with a 1D bug, when in reality, a point is 0D. Significantly, GPT-4 did not pick up nor corrected our error, suggesting a potential limitation in its conceptual comprehension and logical reasoning capabilities.}}

\medskip
\textit{While extending this experiment to an infinite dimensional space instead of just focusing on the bug, might offer further and better insight, the constraints of our current mathematical understanding render us uncertain about the feasibility of this approach with a Möbius strip. We welcome others in the field who may have a more definitive answer to this question.}

\medskip
Great questions!

1. For a 1D bug, which is essentially a point, the concept of orientation doesn't really apply because a point doesn't have a direction or sides. However, the 1D bug (point) would indeed traverse the entire Möbius strip and return to its starting point after one full loop, showing the single-sided nature of the Möbius strip.

2. When considering an n-dimensional bug as n goes to infinity, we're delving into highly abstract territory. We can't really visualize or even conceptualize what a bug existing in infinite dimensions would look like or how it would behave. However, based on the pattern we've seen with lower dimensions, we might hypothesize that an infinite-dimensional bug would experience a "flip" in its orientation in the infinite dimension after one loop and would return to its original orientation after two loops. But again, this is purely speculative and abstract, and can't be directly experienced or visualized.

\medskip
\textcolor{CadetBlue}{\textbf{The extension to an infinite-dimensional bug is complex and necessitates further contemplation. Are you aware of any theorems that could aid us in refining the response provided in your second bullet point?}}

\medskip
When we talk about "infinite" in the context of mathematics, one concept that comes to mind is Georg Cantor's work on different "sizes" of infinity. Cantor's work shows that not all infinities are equal – for example, there are more real numbers between 0 and 1 than there are integers. This could potentially lead to interesting considerations if we were to imagine our "infinite-dimensional bug" inhabiting a space with different "sizes" of infinity in different dimensions, but this is highly speculative.

Another relevant concept is the idea of limit points and convergence in the context of topology and analysis. This could potentially be useful in refining our understanding of what it means for the bug to traverse the Möbius strip an infinite number of times.

It's worth noting that while these concepts might help us to think about the problem in more abstract terms, they don't directly answer the question of what happens to an infinite-dimensional bug on a Möbius strip. This question is a fascinating one, but it's outside the realm of established mathematical theory as of my last training cut-off in 2021.

To truly explore these questions, you might need to \textbf{delve into areas of mathematics like set theory, topology, and higher-dimensional geometry}. These are complex fields that require a solid foundation in mathematical principles, but they can provide the tools needed to explore these abstract and fascinating concepts.

\medskip
\hl{\textit{\textbf{Manual Annotation:}}

\textit{We observe that the theorems cited by GPT-4 bear only superficial connections to the term "infinite", rather than providing substantial relevance to the problem we aim to conceptualize. This may suggest that GPT-4's comprehension of the underlying concepts is not deeply rooted. Nevertheless, it is noteworthy that GPT-4 can identify and highlight potentially related mathematical areas, thereby offering us a directional lead for further exploration and study of our topic.}}\\

\hline
\end{tabularx}
\caption{
Comprehensive log of the brainstorming conversation with GPT-4 for \textit{Experiment I: Möbius and Bugs}.
}
\end{table*}

\begin{table*}
\hypertarget{Exp 1-2 Table}{}
\centering
\begin{tabularx}{\textwidth}{X}
\hline
\textbf{From Möbius to Klein}\\
\hline

\fontfamily{cmss}\selectfont
\smallskip
\textcolor{CadetBlue}{\textbf{Do you know how the Klein bottle can be constructed from Möbius strips? If so, can you help me to visualize and understand this construction process?}}

\medskip
Absolutely, I can help explain the construction of a Klein bottle using Möbius strips!

A Klein bottle is another non-orientable surface, like the Möbius strip, but it's a bit more complex. It's \textbf{a two-dimensional manifold} that cannot be embedded in three-dimensional space without self-intersecting (crossing itself). However, it \textbf{can be embedded in four dimensions without self-intersecting}.

One way to visualize a Klein bottle in 3D (with self-intersection) is to think of it as being constructed from two Möbius strips. Here are the steps:

1. Start with two Möbius strips.

2. Imagine gluing the edges of these two Möbius strips together. This is where things get a little tricky. If we were in four-dimensional space, we could do this without the strips intersecting each other. But in three dimensions, we have to allow the strips to intersect.

3. The resulting object is a Klein bottle. It has the property that if you start at any point on the surface and move in any direction, you will eventually return to your starting point, having covered the entire surface.

Another interesting property of the Klein bottle is that it has no inside or outside. If you were a 2D creature living on the surface of a Klein bottle, you could move from the "inside" to the "outside" without crossing an edge, just like you can move from one "side" to the other of a Möbius strip without crossing an edge.

However, keep in mind that this is a simplified explanation. The actual construction of a Klein bottle is a bit more complex and requires some understanding of topology and higher-dimensional geometry.

I hope this explanation helps give you a basic understanding of how a Klein bottle is related to a Möbius strip. If you want to dive deeper into this topic, feel free to ask!

\medskip
\textcolor{CadetBlue}{\textbf{Of course! Can you elaborate on "If we were in four-dimensional space, we could do this without the strips intersecting each other. But in three dimensions, we have to allow the strips to intersect" using visual methods?}}

\medskip
\hl{\textit{\textbf{Manual Annotation:}}

\textit{Here, to help us foster an intuitive grasp of why a Klein bottle is intersection-free in 4D while being self-intersecting in 3D, GPT-4 skillfully employs an analogy by first guiding us through a simpler transition from 2D to 3D.}}

\medskip
In a four-dimensional space, it's possible to \textbf{"move" parts of the object through that extra dimension to avoid intersecting with itself, in a way we can't in three dimensions}. The limitation of our 3D world forces the Klein bottle to intersect itself when we try to visualize it.

Let's \textbf{simplify and consider a situation in two dimensions first}. Imagine you have two lines on a piece of paper (a 2D world). If you wanted to move one line "over" the other without them crossing, you couldn't do it in this 2D world. However, if you introduce an extra dimension (lift your pencil off the paper and move it over the other line, then put it back down), you can accomplish this. This is essentially what we're doing with the Klein bottle in 4D space.

Now to the Klein bottle. A Klein bottle can be parameterized in 3D space using the following equations:

$$x = (R + r\times cos(a/2)\times sin(b) - sin(a/2)\times sin(2\times b)) \times cos(a)$$
$$y = (R + r\times cos(a/2)\times sin(b) - sin(a/2)\times sin(2\times b)) \times sin(a)$$
$$z = sin(a/2)\times sin(b) + cos(a/2)\times sin(2\times b)$$\\

\end{tabularx}
\end{table*}

\begin{table*}
\centering
\begin{tabularx}{\textwidth}{X}

\fontfamily{cmss}\selectfont
where $a$ and $b$ are parameters that vary over the range $[0, \pi]$, $R$ is the distance from the hole in the middle to the center of the tube, and $r$ is the radius of the tube. These formulas create a 3D representation of a Klein bottle, but it intersects itself at the "neck" where the bottle loops back into itself.

In 4D, there's an additional term that gives the Klein bottle the room it needs to avoid self-intersection:

$$x = (R + r\times cos(a/2)\times sin(b) - sin(a/2)\times sin(2\times b)) \times cos(a)$$
$$y = (R + r\times cos(a/2)\times sin(b) - sin(a/2)\times sin(2\times b)) \times sin(a)$$
$$z = sin(a/2)\times sin(b) + cos(a/2)\times sin(2\times b)$$
$$cos(a/2)\times sin(b) - sin(a/2)\times sin(2\times b)$$

Notice the new $w$ term. This $w$ coordinate is the fourth dimension, which is simply an extra mathematical dimension, in which the Klein bottle can exist without intersection.

Unfortunately, I can't directly illustrate 4D objects here. The equations I've provided represent a mathematical description of the Klein bottle in both 3D and 4D space, and would require software capable of rendering 4D graphics to visualize. The best we can do to visually represent a Klein bottle is to use a 3D projection, but this projection will always include a self-intersection.\\
\end{tabularx}

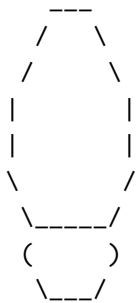
\begin{verbatim}

        ___
       /   \
      /     \ 
     |       |
     |       |
     \       /
      \_____/
      (     )
       \___/
\end{verbatim}

\begin{tabularx}{\textwidth}{X}
\fontfamily{cmss}\selectfont
\medskip
The "neck" of the bottle extends out from the side, loops around and enters through the base, intersecting with the side of the bottle. The intersection point is an artifact of trying to fit the Klein bottle into three dimensions - in four dimensions, this intersection wouldn't occur. The 2D drawing can't do justice to the 3D projection, but it gives a rough idea.

In a true 4D Klein bottle, a creature moving along the surface could travel from the outside to the inside without crossing an edge, similar to the Möbius strip.

\medskip
\textcolor{CadetBlue}{\textbf{Can you treat the fourth dimension as an extra dimension, like "time", and so you can show me a series of 3D projections for better describing the Klein bottle?}}

\medskip
While time is often used as a fourth dimension in the field of physics, it's not quite appropriate to use time as the fourth dimension in this context. In the context of a four-dimensional Klein bottle, the fourth dimension is a spatial dimension, not a temporal one.

However, your idea of using a sequence of 3D projections to better visualize the Klein bottle is a good one. We could imagine taking "snapshots" of the Klein bottle at different stages of its formation. Let's try to visualize it in stages:

\smallskip

1. Start with a cylinder:\\

\end{tabularx}

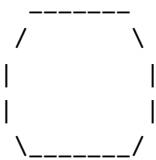
\begin{verbatim}

    _______
   /       \
  |         |
  |         |
   \_______/
\end{verbatim}

\end{table*}

\begin{table*}
\centering
\begin{tabularx}{\textwidth}{X}
2. Now, instead of joining the top and bottom directly, we make a twist similar to a Möbius strip, but we don't connect the ends yet:\\

\end{tabularx}

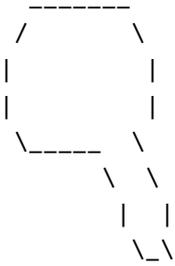
\begin{verbatim}

    _______
   /       \
  |         |
  |         |
   \_____  \
         \  \
          |  |
           \_\
\end{verbatim}

\begin{tabularx}{\textwidth}{X}
\fontfamily{cmss}\selectfont
3. Next, we extend the end through the side of the cylinder:\\

\end{tabularx}

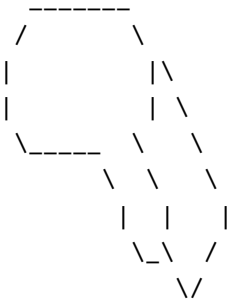
\begin{verbatim}

    _______
   /       \
  |         |\
  |         | \
   \_____  \   \
         \  \   \
          |  |   |
           \_\  /
              \/
\end{verbatim}

\begin{tabularx}{\textwidth}{X}
\fontfamily{cmss}\selectfont
4. Finally, we connect the end back to the bottom. In 3D this would result in a self-intersection, but in 4D it doesn't:\\

\end{tabularx}

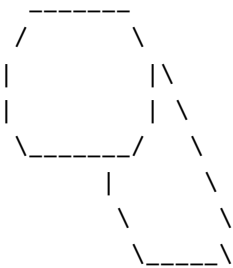
\begin{verbatim}

    _______
   /       \
  |         |\
  |         | \
   \_______/   \
         |      \
          \      \
           \_____\
\end{verbatim}

\begin{tabularx}{\textwidth}{X}
\fontfamily{cmss}\selectfont
The final drawing shows the self-intersection that occurs in 3D. Remember, in a true Klein bottle in 4D, this intersection doesn't happen. This is the best way to visualize the process in stages, but keep in mind this is a very abstract concept and hard to represent fully in three dimensions or less.

\medskip
\textcolor{CadetBlue}{\textbf{Given that we know a Klein bottle can be constructed from two Möbius strips, is it possible to generalize this combination process to using more Möbius strips?}}

\medskip
It's an interesting question! Indeed, the idea of combining topological objects like Möbius strips can lead to other interesting and complex structures. However, when it comes to creating new objects from multiple Möbius strips, the possibilities depend on how the strips are connected.

A Klein bottle can be thought of as being formed from two Möbius strips by identifying (gluing together) their boundaries. This identification results in a non-orientable surface that, unlike a Möbius strip, is also closed (has no boundary).

If we wanted to combine more than two Möbius strips, we would need to think carefully about how to connect their boundaries. For example, three Möbius strips could be connected end-to-end in a cycle, forming a larger non-orientable surface. However, this surface would not be closed and would not be a generalization of the Klein bottle or the Möbius strip.

In another case, if we took four Möbius strips and connected their ends pairwise, we would again form a non-orientable, but now also closed surface. This surface would be more complex than a Klein bottle, having more twists and loops.\\

\end{tabularx}
\end{table*}

\begin{table*}
\centering
\begin{tabularx}{\textwidth}{X}

\fontfamily{cmss}\selectfont
So, while it's certainly possible to combine more than two Möbius strips into a single object, the resulting surfaces depend on the specifics of how the strips are connected. They would be more complex and harder to visualize, especially since we're already dealing with non-orientable surfaces that cannot be fully represented in three dimensions without self-intersection.

It's also worth noting that while the idea of forming a Klein bottle from two Möbius strips can be a useful intuition, \textbf{the formal mathematical definition of a Klein bottle doesn't involve Möbius strips, but rather a certain identification of points on the boundary of a square} (similar to how a Möbius strip can be formed by identifying points on the boundary of a rectangle). This definition generalizes in a different direction, leading to a wide variety of other topological spaces.

\medskip
\textcolor{CadetBlue}{\textbf{Can we formalize "if we took four Möbius strips and connected their ends pairwise, we would again form a non-orientable, but now also closed surface. This surface would be more complex than a Klein bottle, having more twists and loops"?}}

\medskip
\hl{\textit{\textbf{Manual Annotation:}}

\textit{Though not explicitly stated by GPT-4, the Klein bottle, as suggested in the final paragraph, is a type of quotient topological space. Assuming we had no prior knowledge of this, we showcase how GPT-4 can guide our exploratory discussion towards a more formal definition of this particular geometry surface within a mathematical context.}}

\medskip
Sure! In the field of topology, the formal description of complex shapes, like the one formed by joining four Möbius strips, often involves quotient space. A quotient space is formed by taking a standard shape (like a square or a cube) and identifying, or "gluing together", certain points.

\smallskip

To formally define a shape created by joining four Möbius strips, we might proceed as follows:

1. Start with a cube, which we'll consider to be the product of three intervals: $[0,1] \times [0,1] \times [0,1]$.

2. We'll identify points on the boundary of the cube as follows. For points $(x, y, z)$ on the boundary, we'll identify a point with its diametrically opposite point. That is, $(x, y, z)$ is identified with $(1-x, 1-y, 1-z)$.

The resulting space is a quotient space of the cube, and it can be visualized as a sort of four-dimensional analogue of the Klein bottle. The process of forming a quotient space by identifying points can be generalized to create a wide variety of other topological spaces. It is a fundamental technique in the field of topology. This is how we formally define shapes like the Möbius strip and the Klein bottle, and it's the method we would use to formally define a shape created by joining together four Möbius strips.\\

\hline
\end{tabularx}
\caption{
Comprehensive log of the extension to the brainstorming conversation with GPT-4 for \textit{Experiment I: From Möbius to Klein}.
}
\end{table*}

\begin{table*}
\hypertarget{Exp 2 Table}{}
\centering
\begin{tabularx}{\textwidth}{X}
\hline
\textbf{Cats and Dogs}\\
\hline

\fontfamily{cmss}\selectfont
\smallskip
\textcolor{CadetBlue}{\textbf{I will supply you with four 768-dimensional vectors, each of them is a vector representation of a distinct image produced by an image encoder. Can you help find connections among pairs of those vectors?}}

\medskip
Absolutely, I'd be happy to help! In order to analyze these vectors, we will need to perform a few steps:

\smallskip

1. \textbf{Pairwise comparison:} We'll start by calculating the cosine similarity between each pair of vectors. This will provide us with a quantitative measure of how similar each pair of images is based on their vector representations.

2. \textbf{Dimension Reduction:} To visualize these high-dimensional vectors, we can use dimensionality reduction techniques like Principal Component Analysis (PCA) or t-Distributed Stochastic Neighbor Embedding (t-SNE). These techniques will allow us to represent these 768-dimensional vectors in a 2D or 3D space.

3. \textbf{Visualization:} Once we have our reduced dimensions, we can plot these vectors to visually observe any clusters or outliers.

4. \textbf{Analysis:} Based on the quantitative (cosine similarity) and qualitative (visual) findings, we can begin to deduce insights about the relationships between these images.

\smallskip

Let's start with step 1, the pairwise comparison. Can you provide the four vectors?

\medskip
\hl{\textit{\textbf{Manual Annotation:}}

\textit{We have left out the detailed long input vectors, such as $[-1.8285e-02, ..., 4.5278e-04]$, from this log. Each of these vectors corresponds to a specific image - "standing-cat", "sleeping-cat", "standing-dog", and "sleeping-dog" - named according to the actual contents depicted in the original images.
GPT-4 has suggested some promising steps for analyzing the vectors in its above response. However, it appears to lose track of these proposals after processing a chain of input vector sequences.}}

\medskip
\textcolor{CadetBlue}{\textbf{Do you spot any relationship between the vector pairs "standing-cat" and "sleeping-cat"? How about the vector pairs "standing-dog" and "sleeping-dog"? Can you find any correlation in the co-movements of those two pairs? You are welcome to use distance functions such as cosine-similarity. You may also try projecting the original vectors into a different vector space.}}

\medskip
I see you're asking a question that involves word embeddings or word vectors, common in natural language processing (NLP). These word vectors represent words or phrases in multi-dimensional space, and their geometric relationship can indicate semantic relationships between words. However, as an AI text model, I do not have real-time access to any word embedding models such as Word2Vec, GloVe, or FastText, and I can't calculate these vectors or their relationships directly.

That being said, I can still provide a general explanation based on the principles of how these models usually work.

\smallskip

Given the vector pairs "standing-cat" and "sleeping-cat", and "standing-dog" and "sleeping-dog", the vectors "standing-cat" and "standing-dog" may be seen as analogous to "sleeping-cat" and "sleeping-dog", respectively. This is because the word pairs ("standing", "sleeping") and ("cat", "dog") might have similar relationships in the word vector space. For instance, "standing" and "sleeping" might both be considered states or conditions, and "cat" and "dog" are both types of pets. If the word vector model has been trained effectively, it could capture these relationships.

Cosine similarity could be used to measure the angle between these vectors, providing a measure of how similar the vectors are irrespective of their magnitude. If the cosine similarity between "standing-cat" and "sleeping-cat" is close to the cosine similarity between "standing-dog" and "sleeping-dog", it could be interpreted that the state transitions (from standing to sleeping) for cats and dogs are semantically similar.\\

\end{tabularx}
\end{table*}

\begin{table*}
\centering
\begin{tabularx}{\textwidth}{X}

\fontfamily{cmss}\selectfont
About projection to a different vector space, dimensionality reduction techniques like PCA (Principal Component Analysis) or t-SNE could be used. These techniques can help visualize the word vectors by projecting the high-dimensional vectors to 2D or 3D space. However, it's crucial to note that dimensionality reduction might cause some loss of information.

Please note that these interpretations are highly dependent on the corpus the word vectors are trained on. Different corpora could result in different semantic relationships being captured. Also, as I mentioned earlier, I'm providing this information based on principles and theory, as I cannot perform these calculations directly.

\medskip
\textcolor{CadetBlue}{\textbf{Given that "it's crucial to note that dimensionality reduction might cause some loss of information," could you assist in determining the optimal number of dimensions for studying their pairwise relationships? Note that the optimal number of dimensions might exceed 768, their original dimension.}}

\medskip
\hl{\textit{\textbf{Manual Annotation:}}

\textit{The version of GPT-4 we experimented with unfortunately could not directly do numerical computations nor executing a model or piece of code, which we believe, will be accompanying features in a later GPT.
But if it could, our original plan was to do a more data science-oriented approach of exploring and finding the structures directly in the real data we provided.
Here because of the aforementioned limitations, we instead delved into a more conceptual statistical discussion, which also bears its own value.}}

\medskip
As an AI language model, I don't have the ability to directly perform calculations or numerical analysis. However, I can certainly explain the process one would go through to determine the best number of dimensions.

Dimensionality reduction is a complex process that involves making trade-offs between simplicity and information retention. More dimensions generally mean more accurate representations of the data, but also more complexity and potential for overfitting. Conversely, fewer dimensions can lead to simpler models and less overfitting, but at the risk of losing important information.

There isn't a definitive method to determine the best number of dimensions, but there are techniques that can be helpful:

\smallskip

1. \textbf{Scree Plot}: This technique is often used in Principal Component Analysis (PCA). The variance explained by each principal component is plotted, usually in decreasing order. The best number of dimensions is often selected at the "elbow point", where the plot stops decreasing sharply and begins to flatten out. This indicates that adding more dimensions beyond this point will not add much explanatory value.

2. \textbf{Cumulative Explained Variance}: This is another technique used in PCA. A certain percentage (e.g., 95\%) of the total variance is targeted, and dimensions are added until this percentage is reached.

3. \textbf{Cross-validation}: This technique involves building a predictive model using the reduced-dimension data, then evaluating the model's performance. The best number of dimensions is typically the one that results in the best model performance, based on a selected evaluation metric.

\smallskip

Remember, choosing the number of dimensions is a balance between retaining information (variance in the data) and avoiding over-complexity or overfitting. The optimal number may indeed be greater than the original 768 dimensions, especially if the dataset is very complex and high dimensional. However, in practice, it's rare to increase the number of dimensions in dimensionality reduction, as it contradicts the primary goal of the process: reducing complexity.

\medskip
\textcolor{CadetBlue}{\textbf{While I concur that "dimensionality reduction" could be beneficial, I would prefer not to confine myself solely to the "reduction" approach. Rather, my objective is to ascertain the optimal dimensionality of a space to enhance the likelihood of discerning pairwise relationships among the four projected vectors. Could you revise your suggestions to align with this goal?}}\\

\end{tabularx}
\end{table*}

\begin{table*}
\centering
\begin{tabularx}{\textwidth}{X}

\fontfamily{cmss}\selectfont
Given the new information you've provided, it sounds like you're interested in conducting some form of multivariate analysis to understand the relationships between these vectors. If the goal is to identify the optimal number of dimensions for these analyses, there are a few steps you can consider:

\smallskip

1. \textbf{Exploratory Data Analysis (EDA)}: Conduct preliminary analyses to understand the structure and characteristics of your data. This could include calculating basic statistics (mean, variance, etc.), looking at distribution characteristics, or visualizing pairwise relationships with scatter plots.

2. \textbf{Correlation Analysis}: Calculate the correlation matrix for your dataset. This will help you understand the pairwise relationships between each pair of vectors. High correlation between two vectors may suggest they can be represented in a lower-dimensional space.

3. \textbf{Factor Analysis or Principal Component Analysis (PCA)}: Both methods can help in understanding the underlying structure of your data. These techniques will give you a new set of uncorrelated variables, each a linear combination of the original variables (vectors). The Eigenvalues of these new variables can be analyzed to identify the number of significant factors or components, which can hint at the optimal number of dimensions.

4. \textbf{Cross-Validation}: If you're planning on using these vectors in a predictive model, perform cross-validation. Train models with different numbers of dimensions and see which performs the best according to your chosen performance metric.

5. \textbf{Dimensionality Increase}: Rarely, if the complexity of relationships is high, you may consider adding more dimensions (higher than 768) through techniques such as Kernel PCA or using autoencoders in deep learning.

\smallskip

Remember, while PCA and Factor Analysis are powerful tools for understanding data structure, they're based on specific assumptions about the data, and care should be taken to ensure these assumptions are met.

Finally, this process involves a degree of trial and error. It's unlikely that there will be one clear "optimal" number of dimensions that is immediately apparent. Instead, it's a balance of capturing the necessary complexity to understand your data while avoiding overfitting or creating overly complex models. It's possible that different methods will suggest different numbers of dimensions, and you'll need to use your judgment to decide which is the most suitable for your specific purpose.

\medskip
\textcolor{CadetBlue}{\textbf{Using the vectors I provided as an example, could you elucidate on how to employ factor analysis or kernel PCA to identify an optimal number of dimensions to which the vectors should be transformed? Additionally, could you elaborate on the "specific assumptions about the data" when "care should be taken to ensure these assumptions are met"?}}

\medskip
Certainly, I'd be glad to elaborate.

\smallskip

1. \textbf{Factor Analysis}: Factor Analysis is a statistical method used to describe variability among observed, correlated variables in terms of potentially lower number of unobserved variables called factors. The observed variables are modelled as linear combinations of the potential factors, plus some error terms.

For your case, you would first compute a correlation matrix of your vectors. The correlation matrix represents the pairwise relationship among the vectors. Next, you'd perform an Eigenvalue decomposition of the correlation matrix. Each Eigenvalue represents the amount of variance in the data that is accounted for by its corresponding Eigenvector (factor).

The number of factors to retain can be guided by a few rules-of-thumb. You might keep all factors with Eigenvalues greater than 1, which explains more variance than a single original variable. You might also consider a scree plot, where Eigenvalues are plotted in descending order, and you retain factors up to the point where the decline of Eigenvalues appears to level off.

The retained factors can be interpreted by examining the factor loadings, which indicate how strongly each original variable is associated with each factor. This will give you insights into which dimensions (factors) are most significant in describing the pairwise relationships in your data.\\

\end{tabularx}
\end{table*}

\begin{table*}
\centering
\begin{tabularx}{\textwidth}{X}

\fontfamily{cmss}\selectfont
2. \textbf{Kernel PCA}: Kernel PCA is a version of Principal Component Analysis (PCA) which operates in a high-dimensional feature space using what's known as the "kernel trick". By applying a nonlinear mapping to a higher dimension, Kernel PCA allows for the possibility of making complex relationships more visible and easier to work with.
For your data, this means you could apply a Kernel PCA with different kernel functions (e.g., linear, polynomial, Gaussian, sigmoid, etc.) and project your data into higher-dimensional space. This new space may reveal relationships between your vectors that were not visible in the original space.

\smallskip

The optimal number of dimensions can be determined similarly to the way it is in regular PCA. That is, you retain the top principal components that explain the most variance until you reach a suitable threshold (e.g., 95\% of the total variance).

\smallskip

As for the assumptions of these methods:

- Factor analysis assumes that there are linear relationships among your variables (vectors). It also assumes that any noise (errors) in your data is random and normally distributed. If these assumptions aren't met, the factors you derive may not be valid or meaningful.

- Kernel PCA makes fewer assumptions about the underlying data, which is why it can be useful when dealing with complex, nonlinear relationships. However, it does assume that a nonlinear transformation to a higher-dimensional space will make the data more tractable, which may not always be the case.

\smallskip

I hope this provides a useful starting point for exploring these methods with your data!

\medskip
\textcolor{CadetBlue}{\textbf{Considering that my data comprises vector embeddings of images, what kernel function would you recommend as a starting point?}}

\medskip
When choosing a kernel function in Kernel PCA for image data, it often depends on the specific characteristics and properties of your dataset, as well as the problem you're trying to solve. Here are a few kernels that are commonly used:

\smallskip

1. \textbf{Linear Kernel}: If the data is linearly separable (which might not be the case for complex image data), a linear kernel can be used.

2. \textbf{Polynomial Kernel}: This kernel is often used in image processing. It provides a good fit for data where the relationship between the vectors is of a certain degree of polynomial. You could experiment with different polynomial degrees to see what works best for your data.

3. \textbf{Radial Basis Function (RBF) or Gaussian Kernel}: This is one of the most commonly used kernels and is especially effective for high-dimensional data. It can handle non-linear and complex transformations well, which could be suitable for image data. It also has only one parameter to tune (the Gaussian radius), which simplifies things.

4. \textbf{Sigmoid Kernel}: This is used as a proxy for neural networks. However, in practice, it tends not to perform as well as the RBF kernel for many types of data.

\smallskip

\textbf{Given that your data is based on image embeddings and considering the complexity of image data, you might want to start with the RBF kernel as it can handle non-linear and complex transformations}. However, it's generally a good idea to experiment with different kernels and parameters to see what works best for your specific case.

\smallskip

\textbf{\textit{Note:} It's also important to normalize your data before applying Kernel PCA to ensure that all dimensions contribute equally to the distance metric.}

\smallskip

In terms of finding the optimal number of dimensions to project into, \textbf{a good rule of thumb is to look for the point where adding another dimension doesn't significantly increase the variance explained by the model}. This is often visualized as an "elbow" in a plot of explained variance vs number of dimensions. This approach attempts to balance between model complexity (more dimensions) and explained variance.\\

\hline
\end{tabularx}
\caption{Comprehensive log of the brainstorming conversation with GPT-4 for \textit{Experiment II: Cats and Dogs}.}
\end{table*}

\begin{table*}
\hypertarget{Exp 3 Table}{}
\centering
\begin{tabularx}{\textwidth}{X}
\hline
\textbf{A Virtual Conference of the \textit{n}-body problem}\\
\hline

\fontfamily{cmss}\selectfont
\smallskip
\textcolor{CadetBlue}{\textbf{Could you provide me with a brief overview of the n-body problem, along with its latest updates, in layman's terms?}}

\medskip
\hl{\textit{\textbf{Manual Annotation:}}

\textit{We start the conversation by establishing a common understanding between us. Subsequently, we invite GPT-4 to shed light on the challenges intrinsic to this open research problem.}}

\medskip
Sure, I can provide an overview of the n-body problem in layman's terms, though keep in mind my training data only goes up until September 2021, so I may not have the very latest developments. However, I can tell you the current state of understanding up to that point.

The n-body problem is a classic problem in physics and mathematics. It's essentially about predicting the individual motions of a group of celestial objects interacting with each other gravitationally. The "n" in "n-body" stands for any number - so you can have a 2-body problem, a 3-body problem, or a problem with any number of bodies.

For the 2-body problem, the solution is relatively straightforward - for example, the way the Earth orbits the Sun can be very accurately modeled by considering just these two bodies (ignoring the influence of other planets for simplicity). The solution for this case is well-known and leads to elliptical orbits as described by Kepler's laws.

However, as we increase the number of bodies involved, things get increasingly complex. The 3-body problem is a famous example. There's no general solution for arbitrary initial conditions in terms of simple mathematical functions, although some special solutions are known.

The complexity grows rapidly with each added body (4-body problem, 5-body problem, and so on), and for these larger "n-body" problems, solutions are generally only possible through numerical methods - which involve using computers to perform a series of approximations to get an answer.

As of my knowledge cut-off in 2021, there's been much research into special cases and approximation methods, with significant insights coming from fields like chaos theory (which deals with systems that are very sensitive to initial conditions). However, a general, analytical solution to the n-body problem for $n > 2$ remains elusive.

To learn about the latest advances beyond this, you might want to consider looking for recent academic papers, or reliable physics and mathematics news sources. But remember, the n-body problem is a complex and challenging problem that has been puzzling scientists for centuries!

\medskip
\textcolor{CadetBlue}{\textbf{Instead of the general n-body problem, let's talk more about the analytical solutions available for the 3-body problem. Can you summarize the 3-body problem and its solutions in a straightforward and clear manner?}}

\medskip
As of my last training data in September 2021, there isn't a general analytical solution to the three-body problem, only specific solutions under particular conditions. This is a complex topic in physics, but I'll try to summarize the problem and some known solutions as straightforwardly as possible.

The three-body problem originates from celestial mechanics, a field of study in physics. Imagine three celestial bodies, like stars or planets, each with a specific mass. These bodies exert a gravitational pull on each other, leading to a system of motion. The question is, given the initial positions and velocities of these bodies, can we predict their motion at any future time? This is the essence of the three-body problem.

Now, as I mentioned, there's no known general analytical solution to the three-body problem. This means that we can't write down a simple equation or set of equations to solve the problem in all cases. However, we do have some specific solutions for special cases:

\smallskip

1. \textbf{Lagrange's Solution}: In 1772, Joseph-Louis Lagrange discovered a special solution. He showed that if three bodies of equal mass are located at the vertices of an equilateral triangle, they will rotate in a circular orbit around the center of mass while maintaining the equilateral triangle configuration.\\

\end{tabularx}
\end{table*}

\begin{table*}
\centering
\begin{tabularx}{\textwidth}{X}

\fontfamily{cmss}\selectfont
2. \textbf{Euler's Collinear Solution}: Another special case was found by Leonhard Euler, where the three bodies lie on a straight line at all times, oscillating back and forth around the center of mass.

3. \textbf{Restricted Three-Body Problem}: The Restricted Three-Body Problem is another special case. Here, one body is assumed to have negligible mass and doesn't affect the motion of the other two bodies. This can approximate the motion of a spacecraft in the gravitational field of Earth and the Moon, for instance. The well-known Lagrangian points are solutions to this problem.

4. \textbf{Choreographies}: A relatively recent discovery (1993 by physicist Cris Moore) is the existence of solutions where the bodies move along the same closed curve or "choreography". These are complex solutions and the bodies need to be carefully arranged at the start to follow these trajectories.

I must stress that all these solutions are specific and don't provide a general solution to the three-body problem. In most scenarios, numerical methods are used for approximations. The three-body problem remains a central problem in celestial mechanics and dynamical systems, and research into this problem is ongoing.

\medskip
\textcolor{CadetBlue}{\textbf{Leveraging any scientific knowledge you know up to now, can you highlight the primary challenges in attempting to solve the three-body problem analytically?}}

\medskip
Certainly! The three-body problem has fascinated and frustrated mathematicians and physicists for centuries. Here are a few of the primary challenges associated with finding a general, analytical solution:

\smallskip

1. \textbf{Non-linearity of Equations}: The equations governing the motion in the three-body problem are nonlinear. This means that they can't be simplified into a set of linear equations that can be easily solved. This complexity leads to chaotic behavior in the system, where small changes in initial conditions can lead to drastic differences in outcomes.

2. \textbf{Complexity of Gravitational Interactions}: Each body in the system is influenced by the gravity of the other two bodies, and these influences change as the bodies move. This makes for a complex, constantly changing system that's difficult to capture in a single, simple mathematical form.

3. \textbf{Lack of Symmetry}: Unlike the two-body problem, where the bodies' paths are symmetrical, the three-body problem lacks such symmetry. This makes it harder to simplify the problem or find patterns that could lead to an analytical solution.

4. \textbf{Existence of Solutions}: It's unclear whether general solutions even exist for all initial conditions in the three-body problem. For some specific initial conditions, the trajectories of the bodies can become unbounded, with one or more bodies escaping the system entirely.

5. \textbf{The Poincaré Recurrence Theorem}: This theorem, named after Henri Poincaré, suggests that the problem is more complex than we may even be capable of solving. It says that certain systems will, after a sufficiently long but finite time, return to a state very close to the initial state. This means the problem may not have a solution that can be expressed in a finite form.

\smallskip

These challenges have so far prevented scientists from finding a general, analytical solution to the three-body problem. However, the problem has yielded rich insights into the behavior of gravitational systems and has spurred the development of new mathematical techniques and branches of study, such as chaos theory and numerical analysis. Despite its challenges, the three-body problem continues to be an active area of research and a source of fascinating findings.

\medskip
\textcolor{CadetBlue}{\textbf{Suppose you could bring in any relevant mathematicians and scientists from history, introducing them to later discoveries regarding the 3-body problem, and then asking them to contemplate solutions for the challenges you have highlighted. From their discussion, let's collectively attempt to devise a new, potentially viable approach to this problem. Note that everyone is encouraged to pose questions and build upon the ideas of others.}}\\

\end{tabularx}
\end{table*}

\begin{table*}
\centering
\begin{tabularx}{\textwidth}{X}

\fontfamily{cmss}\selectfont
\hl{\textit{\textbf{Manual Annotation:}}

\textit{Here, we introduce and adopt the strategy of engaging great minds and experts from history and a variety of relevant fields for collective brainstorming. Throughout this conversation, we primarily utilize this mechanism, guided by our own intuition and knowledge, to devise a promising new approach to the n-body problem.}}

\medskip
That's a fun and creative scenario! Let's imagine a conference featuring Sir Isaac Newton (formulator of classical mechanics and gravitational theory), Henri Poincaré (pioneer of the study of dynamical systems and chaos theory), and Joseph-Louis Lagrange (one of the key figures in the development of celestial mechanics). Here's how that conversation might go:

\smallskip

\textbf{Newton}: As you may recall, I formulated the laws of motion and universal gravitation which not only explain why apples fall from trees, but also how the moon moves around the Earth, and the planets around the sun. But these laws become quite complicated when we try to apply them to three bodies instead of just two. The gravitational forces all interact with one another, leading to a mess of a situation.

\textbf{Lagrange}: Indeed, Sir Isaac. I found some special solutions where the positions of the bodies form either an equilateral triangle or lie on a straight line, but these cases are very limited. Most three-body systems don't behave in such a neat and orderly fashion.

\textbf{Poincaré}: It seems that the key problem here is the sensitivity to initial conditions - what I've been exploring in my work on dynamical systems. A tiny change in the starting configuration can result in vastly different outcomes. This "chaos" is at the heart of the three-body problem's difficulty.

\textbf{Newton}: That's a valuable insight, Henri. This sensitivity might mean that the problem can't be solved in the conventional sense. Perhaps we need to approach it from a different angle, \textbf{focusing on statistical or probabilistic solutions} instead.

\textbf{Lagrange}: I agree with you, Isaac. And to make matters worse, the equations governing the motion are nonlinear, as you pointed out earlier. This nonlinearity is a significant hurdle.

\textbf{Poincaré}: Instead of trying to find exact solutions, perhaps we should be looking for \textbf{approximate solutions}. We could \textbf{use iterative methods to refine these approximations over time}. This might give us "good enough" solutions for practical purposes, even if we can't find a perfect, closed-form solution.

\textbf{Newton}: That's a promising approach, Henri. Even in my day, we were using numerical methods to make predictions about celestial motion. Today's computational resources would make this approach much more feasible.

\textbf{Lagrange}: Perhaps we can also gain insights from other fields. \textbf{The field of topology, which has emerged after my time, might provide new ways of looking at the problem. For example, we could explore topological features of the solution space.}

\textbf{Poincaré}: I agree, Joseph-Louis. By bringing together different branches of mathematics and physics, and making use of modern computational power, we might make real progress on this ancient problem.

\smallskip

This imaginary conversation represents an ongoing dialogue between scientists over the centuries. The problem hasn't been completely solved yet, but each generation brings new tools and perspectives to bear on the problem, making incremental progress and occasionally achieving breakthrough insights. Even today, scientists are using modern techniques such as machine learning and artificial intelligence to gain new insights into this complex problem.

\medskip
\textcolor{CadetBlue}{\textbf{While the idea of finding approximate solutions is appealing, this method has been exploited to a great extent. Instead, let's shift our focus to exploring the potential existence of a usable analytical solution for "good" initial conditions. I am drawn to Lagrange's suggestion of leveraging findings from the field of topology, so let's guide our discussion in that direction. And as pointed out, we might also employ modern deep learning technologies. Let's extend our conversation with these additional considerations in mind. Feel free to invite other minds who could be helpful!}}\\

\end{tabularx}
\end{table*}

\begin{table*}
\centering
\begin{tabularx}{\textwidth}{X}

\fontfamily{cmss}\selectfont
That sounds like a promising direction! Let's bring into our hypothetical conference a couple more minds: Leonhard Euler (who contributed significantly to various fields, including topology), and Geoff Hinton (a pioneer in deep learning). Let's see how this new conversation might go:

\smallskip

\textbf{Euler}: Lagrange, you mentioned the field of topology, which deals with properties of space that are preserved under continuous transformations. One of the central ideas in topology is the concept of a manifold, a space that is locally similar to Euclidean space. \textbf{Could the space of solutions to the three-body problem be some form of manifold?}

\textbf{Lagrange}: That's an intriguing idea, Leonhard. If we could model the solution space as a manifold, we might be able to use the tools of differential geometry to analyze it. But \textbf{this manifold would exist in a high-dimensional space, since each body's position and velocity contributes three dimensions to the system}. Visualizing or working with such a manifold could be quite challenging.

\medskip
\hl{\textit{\textbf{Manual Annotation:}}

\textit{The three-body problem involves a total of $(3+3)\times 3 = 18$ dimensions, which is quite high for us humans, but not particularly challenging for AI. In the more general $n$-body problem, where $n$ is large, the number of dimensions can be much bigger.}}

\medskip
\textbf{Hinton}: That's where modern technology might come into play. In deep learning, we routinely work with high-dimensional spaces and use techniques like dimensional reduction to make them more tractable. Perhaps we could \textbf{train a neural network to learn the dynamics of the three-body problem, and then analyze the learned model to gain insights}.

\textbf{Poincaré}: Geoff, your work with neural networks is fascinating, and perhaps we could use them as a tool for exploring this high-dimensional manifold. But we would \textbf{need to be careful not to mistake the model's predictions for an analytical solution}. \textbf{It might capture the behavior of the three-body problem accurately, but we would still lack an understanding of the underlying mathematical structure.}

\medskip
\hl{\textit{\textbf{Manual Annotation:}}

\textit{We highlight this aspect as understanding the underlying mechanisms is a critical component in mathematical and numerous scientific research fields, just as important as predicting behaviors and outcomes. LLMs indeed offer a valuable opportunity for us gaining insight into how AI formulates ideas and thoughts.}}

\medskip
\textbf{Newton}: Indeed, Henri. Our goal is not just to predict the system's behavior, but to understand why it behaves as it does. But Geoff's idea of using a neural network as a tool for exploration, not just prediction, is worth considering. We could use it to generate a vast number of potential solutions, then analyze these to look for patterns or structures.

\textbf{Euler}: That could indeed be a useful approach, Isaac. By combining the tools of topology and differential geometry with modern computational methods, we might make real progress towards an analytical solution, or at least a deeper understanding of the problem.

\smallskip

This discussion illustrates how different fields can come together to tackle a challenging problem like the three-body problem. Topology, differential geometry, and modern computational methods like deep learning can all provide valuable tools for exploring the problem space. While this doesn't guarantee a general analytical solution, it could lead to significant insights and new ways of approaching the problem.

\medskip
\textcolor{CadetBlue}{\textbf{Several intriguing aspects emerge from our discussion so far: 1) "Could the space of solutions to the three-body problem be some form of manifold?" 2) "But this manifold would exist in a high-dimensional space, since each body's position and velocity contributes three dimensions to the system. Visualizing or working with such a manifold could be quite challenging." and 3) "In deep learning, we routinely work with high-dimensional spaces."}}\\

\end{tabularx}
\end{table*}

\begin{table*}
\centering
\begin{tabularx}{\textwidth}{X}

\fontfamily{cmss}\selectfont
\textcolor{CadetBlue}{\textbf{Our next step is to elaborate on these points and explore a more concrete method for utilizing deep learning to identify an analytical solution to the three-body problem, specifically in the context of manifolds. To aid us in this endeavor, let's incorporate more expert opinions into the discussion.}}

\medskip
Indeed, those are thought-provoking ideas. Let's continue this hypothetical discussion by bringing in Bernhard Riemann, a mathematician who made significant contributions to the understanding of manifolds and geometry, and Yann LeCun, a leading researcher in machine learning, specifically convolutional neural networks and dimensionality reduction techniques. Here's how they might contribute to the conversation:

\smallskip

\textbf{Riemann}: Euler, your suggestion of treating the space of solutions to the three-body problem as a manifold is intriguing. As we know, a manifold is a space that, in a small enough region, resembles Euclidean space. Since each body's position and velocity contribute three dimensions each, our manifold would exist in a high-dimensional space. The challenge is finding a meaningful representation of this manifold that captures the essence of the problem.

\textbf{LeCun}: Bernhard, in the field of deep learning, we frequently deal with high-dimensional data. A common approach to make these high-dimensional spaces more manageable is through dimensionality reduction. We could use an autoencoder, for example, a type of artificial neural network used to learn efficient codings of input data.

\textbf{Riemann}: Interesting, Yann. \textbf{With an autoencoder, we could learn a lower-dimensional representation of the high-dimensional manifold.} This lower-dimensional representation might be easier to work with and analyze.

\textbf{Hinton}: Yes, the lower-dimensional representation could potentially reveal patterns or structures in the solution space that aren't evident in the high-dimensional space. \textbf{And to connect back to Euler's point, these structures might give us insights into the manifold nature of the solution space.}

\medskip
\hl{\textit{\textbf{Manual Annotation:}}

\textit{In our earlier prompts, we emphasized that "everyone is encouraged to pose questions and build upon the ideas of others". Notably, GPT-4 has indeed demonstrated its ability to recall pertinent points from previous conversations and generate fresh insights based on them. This capacity is crucial for brainstorming, particularly when our goal is to draw upon a wide range of expertise through multiple rounds of collaborative and iterative ideation.}}

\medskip
\textbf{Poincaré}: That sounds promising, but we need to remember our goal here: finding an analytical solution. While a neural network could approximate the solution space, it's essentially a numerical method. We need to extract an analytical insight from it.

\textbf{LeCun}: Indeed, Henri. One possibility could be using the neural network as a generator of solutions. We could then analyze the set of generated solutions to look for common structures or patterns. These patterns could potentially be formalized into an analytical expression or set of expressions.

\textbf{Newton}: So, we're suggesting using a neural network to explore the manifold of solutions, then examining the results to identify possible analytical solutions. That seems like a potentially fruitful approach. The key will be in the analysis of the generated solutions and the extraction of meaningful patterns.

\smallskip

In conclusion, it's suggested that an autoencoder could be used to create a lower-dimensional representation of the high-dimensional manifold of solutions. This representation could be easier to visualize and analyze, potentially revealing structures or patterns in the solution space that could lead to an analytical solution. However, this approach would require careful analysis and mathematical formalization of the patterns identified. It's a creative and potentially promising direction, but also a challenging one, that would require significant advances in both mathematical analysis and machine learning techniques.\\

\end{tabularx}
\end{table*}

\begin{table*}
\centering
\begin{tabularx}{\textwidth}{X}

\fontfamily{cmss}\selectfont
\textcolor{CadetBlue}{\textbf{Rather than relying on humans to analyze and identify patterns through a lower-dimensional representation of the high-dimensional manifold, which results in information loss, can we leverage DL to discover hidden structures of the solution in its original high-dimensional space? I think that a deep neural network should indeed hold such potential. With this insight as our basis, let's carry on with our discussion.}}

\medskip
Certainly! This is a great point to bring up. In essence, you're suggesting that we could potentially use deep learning to go directly from the high-dimensional manifold to an analytical solution. This might require some form of symbolic regression, a technique for finding analytical expressions that fit data. Let's bring in another deep learning expert, Yoshua Bengio, a pioneer in the development of deep learning algorithms, to contribute to the discussion.

\smallskip

\textbf{LeCun}: The suggestion of working directly with the high-dimensional manifold instead of reducing it to lower dimensions is a valid one. Given that deep learning models excel at handling high-dimensional data, there is a possibility that a deep neural network could find patterns directly in this space.

\textbf{Bengio}: I agree with Yann. Deep learning models have demonstrated the ability to work effectively with high-dimensional data in various fields. The task of learning from the high-dimensional manifold of the three-body problem could be thought of as learning representations, which is essentially what deep learning does.

\textbf{Hinton}: True, Yoshua. We've seen success with deep learning in areas with high-dimensional data like image and speech recognition. But this problem is a bit different. We're not just categorizing data, we're looking for an analytical solution — an expression or set of expressions that capture the underlying mathematical structure.

\textbf{Bengio}: That's a good point, Geoff. This would likely require \textbf{a form of symbolic regression, which involves finding an analytical expression that fits the data}. With deep learning, this could mean \textbf{training a model to output an expression instead of a single prediction}.

\textbf{LeCun}: That's an interesting idea, Yoshua. Symbolic regression with deep learning is an active area of research, and it's possible that it could be applied to the three-body problem. But there are still many challenges to be overcome, such as the difficulty of representing and manipulating mathematical expressions within a neural network.

\textbf{Newton}: It seems we're on the cusp of a new approach, using deep learning not just to predict the motion of the bodies, but to derive the underlying equations themselves. This would be a significant step forward, although as Yann mentioned, it's not without its challenges.

\smallskip

So, the consensus appears to be that using deep learning to find an analytical solution directly from the high-dimensional manifold is a promising approach, albeit a challenging one. The concept of symbolic regression, finding an analytical expression that fits the data, seems particularly relevant. However, representing and manipulating mathematical expressions within a neural network is a complex task and is an active area of research in the field of deep learning.

\medskip
\textcolor{CadetBlue}{\textbf{If we consider using symbolic regression to address the three-body problem, are there ways to surmount the potential challenges?}}

\medskip
Certainly, let's bring in another expert for this: Hod Lipson, a pioneer in the field of symbolic regression and genetic programming. Here's how he might contribute to the conversation:

\smallskip

\textbf{Lipson}: The idea of symbolic regression is to search the space of mathematical expressions to find a formula that best fits a given dataset. In a sense, it's like traditional regression, but instead of fitting parameters in a predetermined model, we're searching for the model itself.

\textbf{LeCun}: Right, Hod. So, applying this to the three-body problem, we would be trying to find an equation or set of equations that describe the motion of the three bodies.

\textbf{Lipson}: Exactly. This would require a significant amount of data, likely generated by simulating the three-body problem under various conditions. We would then use symbolic regression to search for a mathematical expression that fits this data.\\

\end{tabularx}
\end{table*}

\begin{table*}
\centering
\begin{tabularx}{\textwidth}{X}

\fontfamily{cmss}\selectfont
\textbf{Hinton}: But as we've mentioned, symbolic regression is quite challenging in a deep learning context. Neural networks are not naturally suited to representing and manipulating mathematical expressions.

\textbf{Lipson}: That's correct, Geoff. We've made some progress in this area using techniques like genetic programming, where we use evolutionary algorithms to search the space of mathematical expressions. But integrating this with deep learning is still an open problem.

\textbf{Bengio}: One possible direction could be to use a differentiable programming approach, where we design neural networks to be more amenable to symbolic manipulation. This is a relatively new field, but it has shown promise in similar tasks.

\textbf{Lipson}: Indeed, Yoshua. Differentiable programming could provide a way to bridge the gap between deep learning and symbolic regression.

\textbf{Newton}: The idea of finding the underlying equations directly from the data, rather than inferring them through observation and deduction, is a significant departure from the way we've traditionally done physics. But it seems to hold promise, especially with the advancements in techniques like differentiable programming.

\medskip
\hl{\textit{\textbf{Manual Annotation:}}

\textit{Newton appears to be greatly inspired by this conversation, which is filled with new methodologies and advancements in various fields that were beyond his reach during his era.}}

\medskip
The consensus appears to be that symbolic regression, potentially in conjunction with techniques like genetic programming or differentiable programming, is a promising approach to finding an analytical solution to the three-body problem. However, integrating symbolic regression with deep learning, and in particular designing neural networks that can represent and manipulate mathematical expressions, is a significant challenge.

\medskip
\textcolor{CadetBlue}{\textbf{I have come to understand that tackling this problem using symbolic regression is challenging. Why don't we reintroduce theoretical physics and topology into the discussion? And let's revisit the idea of potentially using a low-dimensional representation of the high-dimensional manifold.}}

\medskip
Absolutely, your suggestion brings us back to the initial idea of modeling the solutions of the three-body problem as a high-dimensional manifold. Let's bring in Edward Witten, a theoretical physicist known for his work on string theory, which often involves high-dimensional spaces and topology. Here's how he might contribute:

\smallskip

\textbf{Witten}: This has been an insightful conversation so far. I agree with the thought of modeling the solution space as a manifold, and exploring it using deep learning. However, \textbf{the potential information loss from reducing its dimensions should be taken into account}. In string theory, we often deal with high-dimensional spaces, and we might have some ideas to contribute.

\textbf{Euler}: Indeed, Edward. In string theory, you often compactify extra dimensions to make the theory more tractable. Perhaps a similar approach could be used here, retaining the full information of the high-dimensional space while making it easier to work with.

\textbf{Witten}: Exactly, Leonhard. Compactification, or similar techniques, could allow us to study the high-dimensional manifold in a way that is more manageable, while preserving the essential features.

\textbf{LeCun}: Edward, your point is very valid. \textbf{To work with high-dimensional data in machine learning, we often use techniques like convolutional layers that exploit the local structure of the data. Could we find a similar way to exploit the local structure of this high-dimensional manifold?}

\textbf{Witten}: That’s an interesting thought, Yann. In the context of the three-body problem, local structure could refer to the interactions between bodies at a particular moment. The dynamics of these local interactions might provide insights into the overall structure of the manifold.

\textbf{Bengio}: We could \textbf{perhaps train a deep learning model to recognize and learn these local structures, using them as building blocks to construct an approximation of the entire manifold}.\\

\end{tabularx}
\end{table*}

\begin{table*}
\centering
\begin{tabularx}{\textwidth}{X}

\fontfamily{cmss}\selectfont
\textbf{Newton}: And once we have this approximation, we could potentially use symbolic regression or another form of analysis to extract an analytical solution.

\textbf{Witten}: Precisely, Isaac. This would be a significant challenge, of course, but it seems like a promising direction. We might be able to apply some of the mathematical techniques developed in string theory to help with this task.

\smallskip

The discussion suggests a potential new approach: combining ideas from string theory and deep learning to study the high-dimensional manifold of solutions to the three-body problem. Techniques like compactification could be used to make the manifold more manageable, while deep learning could help identify and learn local structures within the manifold. These structures could then be used to build an approximation of the manifold, which could potentially be analyzed using symbolic regression or a similar method to extract an analytical solution. However, this is a complex and challenging task that would require significant advances in several areas.

\medskip
\textcolor{CadetBlue}{\textbf{This is indeed a fascinating and thought-provoking approach that I would not have conceived on my own. Let's address any remaining issues and then briefly summarize the setup of the three-body problem along with the method we have developed so far. Additionally, let's outline the type of data we will need to train the DL model and suggest an appropriate collection method. Have we missed any other challenges that need to be addressed?}}

\medskip
Absolutely, let's present a brief potential approach based on our discussion:

\smallskip

\textbf{1) Problem Statement:}

We consider three point masses under the influence of gravitational forces in three-dimensional space, where the solution is a time evolution of their positions. The set of all possible solutions forms a high-dimensional manifold, each point on the manifold representing a specific state of the three-body system.

\smallskip

\textbf{2) Approach:}

We propose using a deep learning model to learn this manifold. The model would be trained on a large dataset generated by simulating the three-body problem under a variety of initial conditions. Techniques from string theory, such as compactification, could be used to make this high-dimensional manifold more manageable, while preserving the essential features.

The DL model would need to identify and learn local structures within the manifold. These structures could then be used as building blocks to construct an approximation of the manifold.

\smallskip

\textbf{3) Data Collection:}

The deep learning model would require a vast amount of simulated data. This data would consist of time evolution of three-body systems under a variety of initial conditions. The creation of such a dataset would be computationally intensive but feasible with modern technology.

\smallskip

\textbf{4) Challenges and Considerations:}

One of the main challenges in this approach is the integration of symbolic regression with deep learning. Representing and manipulating mathematical expressions within a neural network is a complex task and is an active area of research.
Another challenge comes from the chaotic nature of the three-body problem. Small changes in initial conditions can lead to large differences in outcomes, which might make the learning task more difficult.
Lastly, the issue of data sufficiency may arise. As the manifold is of high dimensionality, covering the space adequately with simulations might be a computationally demanding task.

\smallskip

Overall, the combination of deep learning, symbolic regression, and string theory presents a promising new approach to finding an analytical solution to the three-body problem, but a few challenges remain. Continued research and development in these areas will be crucial for successfully carrying out this approach.\\

\hline
\end{tabularx}
\caption{Comprehensive log of the brainstorming conversation with GPT-4 for \textit{Experiment III: The \textit{n}-body problem}.}
\end{table*}

\begin{table*}
\hypertarget{Exp 4 Table}{}
\centering
\begin{tabularx}{\textwidth}{X}
\hline
\textbf{The wicked Queen and the seven Dwarfs}\\
\hline

\fontfamily{cmss}\selectfont
\smallskip
\textcolor{CadetBlue}{\textbf{Let's think about this problem together:
"Seven dwarfs are imprisoned by the wicked queen who puts a red hat or a green hat on the head of each dwarf. The hats are chosen randomly and every configuration is equally likely. The dwarfs can see all the hats except for his own. At a signal, each dwarf can stay silent, or guess the color of his hat. The queen promises to free all seven dwarfs if at least one dwarf guesses his hat correctly and no one guesses his hat wrong. If all the dwarfs are silent, or some dwarfs say an incorrect color, the dwarfs remain captured. Find a strategy for the dwarfs to go free with probability greater than 80\%."
To facilitate our communication, let's first assign integers from 1 to 7 to the seven dwarfs, from the shortest to the tallest. Let's also adopt a binary notation for the two hat colors: 0 for green, and 1 for red. With this notation, each possible configuration of hats can be represented as a 7-dimensional vector, $c=[c_1, c_2, c_3, c_4, c_5, c_6, c_7] \in \{0,1\}^7$. Could we proceed from here using these notations?}}

\medskip
\hl{\textit{\textbf{Manual Annotation:}}

\textit{GPT-4's initial attempt is completely incorrect. It appears to apply a common "binary parity" strategy from similar-looking problems. Its core logic breaks down in the area we have highlighted below. Despite this, it is not a bad attempt, as this is a common mistake made by people as well.}}

\medskip
Your notation makes sense and will help in discussing this problem more clearly. The strategy I would suggest involves using the concept of "parity", which essentially means whether the sum of certain numbers is even or odd. This can be done in the binary system as well.

The dwarfs can agree on the following strategy:

1. Dwarf 1 will be the one who speaks up and takes on the role of the "guesser", and the rest will remain silent.

2. Dwarf 1 will observe the colors of the hats on the other six dwarfs, add up the binary numbers (where red=1, green=0), and determine whether the total is even or odd.

3. Dwarf 1 will then guess the color of his own hat based on whether he needs to make the total count of red hats odd or even. The other six dwarfs will remain silent.

Now, let's examine the possible outcomes of this strategy. There are $2^7 = 128$ possible hat configurations. Since every configuration is equally likely, each has a probability of 1/128.

In half of these configurations (64), the number of red hats will be even. In the other half (64), the number of red hats will be odd. Therefore, the dwarfs have a 50\% chance of being correct if Dwarf 1 is simply guessing the color of his hat randomly.

\textbf{However, by using the above strategy, Dwarf 1 isn't simply guessing; he's making an educated guess based on the parity of the other six dwarfs' hats. If the sum of the colors of the hats he sees is even, he will guess his own hat is green (0). If the sum is odd, he will guess his own hat is red (1). This strategy effectively doubles their chance of correctly guessing, taking it from 50\% to 100\%.}

Therefore, if they follow this strategy, the dwarfs will be able to go free with 100\% certainty, which is definitely greater than 80\%.

\medskip
\textcolor{CadetBlue}{\textbf{In the statement "If the sum of the colors of the hats he sees is even, he will guess his own hat is green (0). If the sum is odd, he will guess his own hat is red (1)", you have assumed that the total parity of all the hats is even, which only has a 50\% chance. So can this strategy genuinely guarantee a 100\% chance of freeing the dwarfs?}}

\medskip
You're correct, and I apologize for the confusion. It's important to highlight that the strategy relies on an agreed-upon rule beforehand among the dwarfs: that the "guesser" (Dwarf 1) will make his guess based on maintaining a specific parity - let's say even for the purpose of this explanation.

In this case, Dwarf 1 would guess his own hat color such that the total number of red hats (counted as 1s) among the seven dwarfs would be even. Now, half of the 128 possible hat configurations would have an even number of red hats, and the other half would have an odd number.\\

\end{tabularx}
\end{table*}

\begin{table*}
\centering
\begin{tabularx}{\textwidth}{X}

\fontfamily{cmss}\selectfont
By guessing his own hat color to maintain the agreed-upon even parity:

1. If the actual total number of red hats among the seven dwarfs is even, and Dwarf 1 observes an even number of red hats among the other six, he would guess his own hat is green (0), keeping the total even.

2. If the actual total number is even, but he observes an odd number of red hats among the other six, he would guess his own hat is red (1), again keeping the total even.

3. If the actual total number is odd, and he observes an odd number of red hats among the other six, he would guess his own hat is green (0), flipping the total to even.

4. If the actual total number is odd, but he observes an even number of red hats among the other six, he would guess his own hat is red (1), again flipping the total to even.

In all cases, Dwarf 1 guesses in such a way as to maintain the even total parity. As a result, he will be correct in exactly half of the cases: specifically, when the actual total number of red hats among all seven dwarfs is even.

So, the strategy doesn't guarantee their freedom with 100\% certainty, but rather increases their chances to 50\%, which is unfortunately less than the 80\% target. I apologize for the previous miscommunication.

We need to find a strategy with a higher success rate. Would you like me to think about a different strategy, or do you already have some thoughts?

\medskip
\hl{\textit{\textbf{Manual Annotation:}}

\textit{GPT-4 appears to be able to recognize its error after an extensive line of reasoning, which is beneficial for our understanding of where its initial thought process was flawed.}}

\medskip
\textcolor{CadetBlue}{\textbf{Let's try a different approach here. Why not you go ahead and give us some ideas to start with?}}

\medskip
We can \textbf{take inspiration from the concept of Hamming codes} in error detection and correction. In coding theory, a Hamming distance between two vectors is the number of positions at which the corresponding values are different. In this case, it's the number of different-colored hats between any two assignments. In this scenario, an "error" would be a dwarf guessing their hat color incorrectly.

We can construct a set of assignments with a minimum Hamming distance of 3 by using a (7,4,3) Hamming code. This is a linear error-correcting code that encodes 4 bits of data into 7 bits by adding three parity check bits.

Here's an example of 16 possible assignments for the dwarfs (the four data bits are first, followed by the three parity bits):

0000 000\\
0001 101\\
0010 110\\
0011 011\\
0100 111\\
0101 010\\
0110 001\\
0111 100\\
1000 011\\
1001 110\\
1010 101\\
1011 000\\
1100 100\\
1101 001\\
1110 010\\
1111 111\\

\end{tabularx}
\end{table*}

\begin{table*}
\centering
\begin{tabularx}{\textwidth}{X}

\fontfamily{cmss}\selectfont
This table gives us 16 possible assignments of hats to the dwarfs such that the Hamming distance between any two of them is at least 3. 

Is there anything you would like to add or any other approach you have in mind?

\medskip
\textcolor{CadetBlue}{\textbf{This seems to be a promising starting point! We can now imagine each such possible assignment in the reference set you just constructed as a 7-dimensional vector. By taking a "ball" of radius 1 centered at each element in the set and call it a 1-ball (take 0000000 for example), can you sketch us its neighboring 1-balls and show that whether any two balls in this set would overlap or not?}}

\medskip
Given the high-dimensional nature of the problem, sketching out a 7-dimensional graph isn't feasible or intuitive. However, I can explain the concept with reference to a lower-dimensional example and then extend it to the 7-dimensional case.

A "1-ball" in this context is a set of all points (in this case, configurations) with a Hamming distance of 1 from a given center point. In a binary system, a Hamming distance of 1 means flipping one bit in the vector.

If we take the vector 0000000 for example, its 1-balls (vectors at a Hamming distance of 1) would be the vectors obtained by flipping each bit once:

1000000\\
0100000\\
0010000\\
0001000\\
0000100\\
0000010\\
0000001\\

Now, considering our constructed set of 16 vectors with a minimum Hamming distance of 3 between any pair, you can see that none of these 1-balls overlap with our original set. This is because all vectors in the constructed set differ from the vector 0000000 by at least a Hamming distance of 3. In other words, you would have to flip at least three bits to reach any vector in our set starting from 0000000, and flipping just one bit (forming the 1-ball) wouldn't be enough.

\medskip
\hl{\textit{\textbf{Manual Annotation:}}

\textit{To be precise, here GPT-4 should mention that flipping one bit on each of two balls, thus a total of 2 bits, is insufficient. This requirement gives rise to the need for a minimum Hamming distance of 3 as opposed to 2.}}

\medskip
This logic can be applied to all vectors in the constructed set. For each vector, you would have to flip at least three bits to reach another vector in the set. Hence, the 1-balls (flipping only one bit) around each vector in our set will never overlap with any other vectors in the set.

So in a more conceptual sense, each of these vectors in our 7-dimensional space has a 1-ball around it that doesn't intersect with any other 1-ball, guaranteeing the minimum Hamming distance of 3 we established with our set.

\medskip
\textcolor{CadetBlue}{\textbf{Cool! Now combine the information that each dwarf can see the colors of hats of the other 6 dwarfs, the dwarf can then derive two possible configurations of all 7 hats, with our discovery of no two 1-balls in the set would overlap in the 7-dimensional space, because the two completed configurations are one distance apart, therefore they cannot both live in the reference set (but one of them may be), can you help think of a strategy?}}

\medskip
Yes, the dwarfs can use the Hamming (7,4,3) code set as a reference and devise a strategy based on this. Here's the proposed strategy:

1. Each dwarf will look at the colors of the hats of the other six dwarfs.

2. Based on this, each dwarf can think of two possible configurations for all seven dwarfs' hats. These two possible configurations differ only by the color of his own hat.

3. Each dwarf then checks whether these two possible configurations exist in the reference set constructed using the Hamming (7,4,3) code. There are three possibilities:\\

\end{tabularx}
\end{table*}

\begin{table*}
\centering
\begin{tabularx}{\textwidth}{X}

\fontfamily{cmss}\selectfont
a. Neither of the two configurations exist in the reference set. In this case, all dwarfs should stay silent because they can't confidently guess the color of their own hat.

b. One of the two configurations exists in the reference set. This can happen only when the actual 7-hat configuration doesn't lie in the reference set. The dwarf who identifies his corresponding configuration in the reference set should guess the color of his own hat such that his guess brings the actual configuration into alignment with the reference configuration. The remaining six dwarfs should stay silent.

c. Both configurations exist in the reference set. This should never happen, because we have ensured that any two vectors in the reference set have a Hamming distance of at least 3, but the two configurations each dwarf can think of are only 1 distance apart.

With this strategy, when the actual 7-hat configuration doesn't lie in the reference set, one of the dwarfs will be able to identify this situation and make a correct guess, freeing all the dwarfs.

Next, let's analyze the probability of success of this strategy. Notice that there are 128 total possible 7-hat configurations, of which 16 lie in the reference set. Therefore, there are 112 configurations that don't lie in the reference set. For each of these 112 configurations, exactly one dwarf will find a match in the reference set and will be able to make a correct guess. Therefore, the probability of success is 112/128 = 87.5\%, which is greater than 80\%.

Thus, this strategy will be successful with a probability greater than 80\%, meeting the goal set by the dwarfs!

\medskip
\hl{\textit{\textbf{Manual Annotation:}}

\textit{Firstly, we are intrigued by GPT-4's ability to derive this complex strategy as a brainstorming partner. While there are several minor issues in its statements, we spotlight a few for further discussion. It is also worth noting that GPT-4's detailed chain-of-thought reasoning presented below assists us in understanding and tracing the possible origin of the missteps.}}

\medskip
\textcolor{CadetBlue}{\textbf{In fact, the statement "One of the two configurations exists in the reference set" does not necessarily imply "the actual 7-hat configuration doesn't reside in the reference set". And in "The dwarf who identifies his corresponding configuration in the reference set should guess the color of his own hat such that his guess brings the actual configuration into alignment with the reference configuration", it seems to me that the dwarf should instead guess the color of his own hat such that his guess falls outside the reference configurations. Could you clarify if my understanding is correct and explain why?}}

\medskip
You're correct, and I apologize for the confusion. The "one of the two configurations exists in the reference set" scenario can indeed occur when the actual 7-hat configuration is already in the reference set. In this case, a dwarf would identify that the other configuration (i.e., the one involving a different color for his own hat) doesn't belong in the reference set. This suggests that the configuration involving the other color for his own hat is less likely to be the correct one.

Let's clarify this by focusing on the 7-dimensional space and non-overlapping 1-balls.

In our 7-dimensional space, we have 16 "balls" (i.e., configurations) from the reference set, each with a 1-ball around it that doesn't intersect with any other 1-ball. Each of these 1-balls includes all configurations that differ from the center configuration by a Hamming distance of 1 (i.e., flipping one dwarf's hat color).

The fact that these 1-balls don't overlap means that for any given configuration in the reference set, the configurations that can be reached by changing one hat color (which a dwarf could think of based on the hats he sees) are not in the reference set. This is the key to our strategy: it guarantees that the dwarfs can never be confused by seeing a configuration that might be the 1-ball of multiple configurations from the reference set.

If the actual configuration is not in the reference set, then there is exactly one dwarf who will see a configuration that is in the reference set. That dwarf will realize that the configuration he sees is in the reference set, and will know that the actual configuration can be reached by changing his own hat color. Since the actual configuration is not in the reference set, he should guess that his own hat is the color that brings the configuration outside of the reference set.\\

\end{tabularx}
\end{table*}

\begin{table*}
\centering
\begin{tabularx}{\textwidth}{X}

\fontfamily{cmss}\selectfont
Thus, the correct strategy for the dwarfs is: if a dwarf sees a configuration in the reference set, he should guess that his own hat is the color that brings the configuration outside of the reference set.

With this strategy, when the actual 7-hat configuration doesn't lie in the reference set, exactly one dwarf will see a configuration in the reference set and will make a correct guess, freeing all the dwarfs.

Thank you for pointing out this important clarification.

\medskip
\textcolor{CadetBlue}{\textbf{While some minor issues persist, I overall really like this strategy! The corrected version further clarifies why, when "Neither of the two configurations exists in the reference set", not "all dwarfs should stay silent because they can’t confidently guess the color of their own hat"; instead, at least one would make a guess. Could you summarize the solution we have derived so far and explain how we arrived at it?}}

\medskip
Absolutely, I'd be happy to summarize our approach. Here's an overview of the strategy and how we came up with it:

We began with the problem of seven dwarfs, each wearing either a red or green hat. They could only see the colors of the other dwarfs' hats, not their own. The dwarfs wanted to devise a strategy to correctly guess at least one hat color without any incorrect guesses, which would free them from the wicked queen's imprisonment.

To tackle this, we adopted binary notation for the two hat colors: 0 for green, and 1 for red. This transformed the problem into one of binary configurations: seven dwarfs with binary-coded hats became a seven-dimensional binary vector. We could then represent all possible hat assignments as points in a 7-dimensional space, each point being a 7-dimensional binary vector.

We then constructed a reference set of 16 hat assignments using a (7,4,3) Hamming code. This code provided a set of 16 seven-bit binary vectors, with the property that any two vectors in the set had a minimum Hamming distance of 3. Essentially, these 16 vectors formed 16 non-overlapping 1-balls in the 7-dimensional space.

Our strategy for the dwarfs was then built upon this reference set and the properties of these non-overlapping 1-balls:

1. Each dwarf looks at the colors of the other six dwarfs' hats, thereby identifying a 6-bit partial configuration.

2. Each dwarf then imagines two complete 7-bit configurations: one with his own hat as 0 (green), and the other with his own hat as 1 (red).

3. Each dwarf checks these two configurations against the reference set. There are three possibilities:

a. Neither configuration is in the reference set. In this case, all dwarfs stay silent, as they cannot confidently guess their own hat color.

b. One configuration is in the reference set. This can only happen when the actual hat configuration is not in the reference set. The dwarf identifying his corresponding configuration in the reference set should guess that his own hat color is the one that is \textbf{not} in the reference set (ensuring that his guess brings the actual configuration outside of the reference set). All other dwarfs remain silent.

c. Both configurations are in the reference set. This should never occur, because it would mean that two points in the reference set have a Hamming distance of 1, which contradicts our construction of the reference set.

With this strategy, when the actual 7-hat configuration doesn't lie in the reference set, exactly one dwarf will be able to make a correct guess, ensuring the dwarfs' freedom. Since 112 out of the 128 total configurations are outside of the reference set, the probability of success is 112/128 = 87.5\%, which is greater than 80\%, meeting the dwarfs' goal.

This strategy represents a fascinating interplay between the ideas of error-correcting codes, binary configurations, and the geometrical concept of non-overlapping balls in high-dimensional space. It also showcases the power of collective reasoning.\\

\hline
\end{tabularx}
\caption{
Comprehensive log of the brainstorming conversation with GPT-4 for \textit{Experiment IV: The wicked Queen and the seven Dwarfs}.
}
\end{table*}

\begin{table*}
\hypertarget{Exp 5 Table}{}
\centering
\begin{tabularx}{\textwidth}{X}
\hline
\textbf{Probability and Logic}\\
\hline

\fontfamily{cmss}\selectfont
\smallskip
\textcolor{CadetBlue}{\textbf{A professor, noticing that his 100 PhD students have been excessively preoccupied with their phones, thereby significantly delaying their publications, decides to confiscate all their devices. He randomly places the 100 phones into 100 separate boxes in a locked room, with the boxes labeled from 1 to 100.}}

\textcolor{CadetBlue}{\textbf{One technologically adept student manages to hack the lock, but the hack comes with a catch: it only permits students to sneak into the room one at a time, and the room will only be accessible for a brief time window. This leaves no room for communication between any two students' turns.}}

\textcolor{CadetBlue}{\textbf{Once inside the room, each student can randomly select a box and check if it contains their phone. They can repeat this process up to 50 times. If they find their phone, they will take it, but the box must be left behind. This means that a subsequent student entering the room wouldn't know if a box has already been checked and might end up wasting a step to examine it. As a consequence of the hack, the room will be permanently locked after all 100 students have had their turn.}}

\textcolor{CadetBlue}{\textbf{As one of these PhD students, you have a burning desire to retrieve your phone. You particularly miss a picture of your adorable bunny which you need for your ongoing research, so recovering your phone is crucial. The others have similar reasons for wanting their phones back. Can you devise a clever strategy ensuring that everyone finds their phone with a success rate exceeding 30\%?}}

\medskip
To maximize the chance of each student finding their phone, they can use a strategy known as "Designated Box Strategy". Here is how it works:

\smallskip
1. Assign each student a unique number from 1 to 100 (which will correspond to their "designated box"), and let the students memorize their numbers before they enter the room. This could be based on their names, student ID numbers, or any other system that ensures each student gets a different number.

2. When a student enters the room, they start by looking in their designated box. If they find their own phone, they're done and they leave the room.

3. If their phone is not in their designated box, they look in the box that corresponds to the number on the phone they found. They repeat this process, always looking in the box that corresponds to the number on the phone they found most recently, until they either find their own phone or have opened 50 boxes, at which point they must leave the room.

\smallskip
This strategy works because it effectively creates 100 distinct cycles among the boxes, one for each student. Each cycle contains a series of boxes that will eventually lead a student back to their own phone, as long as the cycle is not longer than 50 boxes. The problem is effectively transformed into finding a cycle in a permutation.

If you run a simulation of this problem using this strategy, you'll find that each student has a chance of finding their own phone that is roughly 31.18\%, so it satisfies the requirement of a success rate greater than 30\%.

\medskip
\textcolor{CadetBlue}{\textbf{To demonstrate your understanding, could you determine the probability of the students finding their phones through a step-by-step calculation?}}

\medskip
Yes, there is a mathematical way to calculate the success probability for each student. The key is to understand the distribution of cycle lengths in a random permutation of 100 elements.\\

\end{tabularx}
\end{table*}

\begin{table*}
\centering
\begin{tabularx}{\textwidth}{X}

\fontfamily{cmss}\selectfont
In this case, the boxes and phones create a random permutation of 100 elements, and each student follows a cycle within this permutation. The probability that a student will find their phone is the same as the probability that the cycle they follow is 50 or fewer boxes long.

First, let's consider the total number of permutations of 100 elements. This is simply 100 factorial (100!).

Then, for each possible cycle length $n$ from 1 to 50, we'll calculate the number of permutations that have a cycle of length $n$. To form a cycle of length $n$, we first choose $n$ elements from 100 (which can be done in 100 choose $n$ ways), then arrange these $n$ elements into a cycle (which can be done in $(n-1)!$ ways), and finally arrange the remaining $100-n$ elements in any order ($(100-n)!$ ways). So, the total number of permutations with a cycle of length $n$ is:

$$\binom{100}{n} \times (n-1)! \times (100-n)!$$

To get the total number of successful permutations (i.e., permutations where each student finds their phone), we need to sum this quantity over all n from 1 to 50.

Let's denote this sum as S. Then, the probability that a student will find their phone is S divided by the total number of permutations, which is 100!.

To calculate this probability exactly, we would need to compute the quantity $S$ and then divide by 100!. However, this computation is quite involved due to the large numbers involved (100 factorial is a 158-digit number). In practice, it would be easier to use a computer program to do this computation.

\medskip
\textcolor{CadetBlue}{\textbf{Before calculating 100!, could you simplify your equation first to avoid dealing with excessively large numbers?}}

\medskip
You're correct, we can simplify the equation to avoid handling very large numbers directly. The equation for the number of permutations with a cycle of length $n$ is:

$$\binom{100}{n} \times (n-1)! \times (100-n)!$$

which simplifies to:

$$100! / [n \times (100-n)!] \times (n-1)! \times (100-n)!$$

And this simplifies further to:

$$100! / (n \times (100-n)!)$$

When we divide this by 100!, we get the probability of a cycle of length $n$:

$$1 / (n \times (100-n)!)$$

So, the total probability that a student will find their phone (i.e., the probability that they are in a cycle of length 50 or less) is the sum of this quantity for $n$ from 1 to 50:

$$\sum_{n=1}^{50} 1 / (n \times (100-n)!)$$

This is a much simpler computation that can be done using a computer program, but the exact result will still require a precise computation due to the factorial in the denominator.

It's worth noting that this exact calculation will confirm the result found by simulation, that each student has a chance of finding their own phone that is roughly 31.18\%.\\

\end{tabularx}
\end{table*}

\begin{table*}
\centering
\begin{tabularx}{\textwidth}{X}

\fontfamily{cmss}\selectfont
\textcolor{CadetBlue}{\textbf{"$100! / [n \times (100-n)!] \times (n-1)! \times (100-n)!$. And this simplifies further to: $100! / (n \times (100-n)!)$" is not right.}}

\medskip
I apologize for the confusion, and you're correct. The simplification step was incorrect. Let's correct that:

The number of permutations with a cycle of length $n$ is:

$$\binom{100}{n} \times (n-1)! \times (100-n)!$$

which simplifies to:

$$100! / [n \times (100-n)!] \times (n-1)! \times (100-n)!$$

The $(100-n)!$ terms cancel out, leaving us with:

$$(100! / n) \times (n-1)!$$

Dividing by 100! to get the probability gives us:

$$P(n) = (n-1)! / n$$

So, the probability that a student is part of a cycle of length 50 or less is the sum of these probabilities from $n$=1 to $n$=50.

That sum can be computed directly and will yield the desired result. It's also worth noting that the factorial function grows very rapidly, so even though the sum involves terms up to 50!, the total will still be a reasonable number less than 1.

\medskip
\textcolor{CadetBlue}{\textbf{"which simplifies to: $100! / [n \times (100-n)!] \times (n-1)! \times (100-n)!$" is not right.}}

\medskip
I apologize for the confusion, and you're right. The simplification was incorrect in my previous responses. Let's correct that.

The number of permutations with a cycle of length $n$ is:

$$\binom{100}{n} \times (n-1)! \times (100-n)!$$

This simplifies as:

$$[100! / (n! \times (100-n)!)] \times (n-1)! \times (100-n)!$$

The $(n-1)!$ terms cancel out:

$$= 100! / [n \times (100-n)!] \times (100-n)!$$

The $(100-n)!$ terms cancel out as well, giving:

$$= 100! / n$$

Dividing this by the total number of permutations (100!) gives the probability of a permutation having a cycle of length $n$:

$$P(n) = 1/n$$\\

\end{tabularx}
\end{table*}

\begin{table*}
\centering
\begin{tabularx}{\textwidth}{X}

\fontfamily{cmss}\selectfont
Finally, the total probability that a student will find their phone (i.e., the probability that they are in a cycle of length 50 or less) is the sum of these probabilities from $n$=1 to $n$=50.

Therefore,
$$P = \sum_{n=1}^{50} 1/n$$

This is the harmonic number $H_{50}$, which is approximately 4.5. Since this is more than 1, we should realize that we've made a mistake.

The error is due to the fact that we've double-counted many permutations. Each permutation with a cycle of length $n$ was counted once for each possible position of that cycle, which is not correct. To correct this, we need to divide our count of permutations with a cycle of length $n$ by $n$, because a cycle of length $n$ can appear in $n$ different positions.

So, the correct probability of a permutation having a cycle of length $n$ is:

$$P(n) = 1 / (n \times n)$$

And the total probability that a student will find their phone is the sum of these probabilities from $n$=1 to $n$=50:

$$P = \sum_{n=1}^{50} 1/(n \times n)$$

This is the sum of the reciprocals of the first 50 square numbers. This sum converges to $\pi^2/6$, which is approximately 1.64493. So, each student has a chance of finding their own phone that is roughly 1.64493/5 = 0.328986, or about 32.90\%.

I hope this explanation clarifies the issue, and I apologize for the earlier confusion.

\medskip
\textcolor{CadetBlue}{\textbf{The actual issue with "This is the harmonic number $H_{50}$, which is approximately 4.5. Since this is more than 1, we should realize that we've made a mistake." was that the strategy works if and only if all cycles of permutation for all students have a length no greater than 50.}}

\medskip
You're absolutely right. My apologies for the confusion. I misunderstood the premise. The total probability being greater than 1 indicates an error because we can't have a probability of more than 100\%.

For this problem, each student starts from their designated box (based on their unique number) and follows a sequence (cycle) determined by the placement of the phones. The key point is that the cycle each student ends up following needs to have a length of 50 or less for the student to find their phone within the 50 allowed box openings.

The error in the previous computation was that we considered cycles of any length up to 100, but we should only consider cycles of length up to 50, because a student can only open 50 boxes.

So, the correct probability calculation should consider that we are looking for cycles of length 50 or less in a permutation of 100 elements. This is a rather complex problem to solve analytically, as it involves considering the distribution of cycle lengths in random permutations.\\

\end{tabularx}
\end{table*}

\begin{table*}
\centering
\begin{tabularx}{\textwidth}{X}

\fontfamily{cmss}\selectfont
One way to approach it is to consider the probability of a cycle of length n NOT being present. Since cycles of different lengths are not independent (for example, you can't have both a cycle of length 70 and a cycle of length 40 in the same permutation), it's challenging to directly calculate the probability of having a cycle of length less than or equal to 50.

To make it simpler, it's easier to compute the complementary probability - the probability that a permutation has a cycle of length more than 50. Once we have that, we can subtract it from 1 to get the desired probability.

Let's denote this probability as $P(N)$, where $N$ is the total number of elements in the permutation (in this case, 100). We can express $P(N)$ in terms of smaller probabilities:

$$P(N) = 1/N + (N-1)/N \times P(N-1)$$

This is because a permutation of $N$ elements has a cycle of length more than 50 either if the first element is part of such a cycle (which happens with probability $1/N$) or if it isn't but the remaining $N-1$ elements have a cycle of length more than 50 (which happens with probability $(N-1)/N \times P(N-1)$).

This is a recursive relation that we can use to compute $P(N)$ for any $N$. For $N=100$ and considering that $P(N) = 0$ for $N \le 50$, we get $P(100)$ which is the probability that a permutation has a cycle of length more than 50. Subtracting this from 1 will give us the desired probability - the chance that each student will find their phone.

In reality, this computation would be quite challenging to perform by hand, but it could be done easily with the help of a computer program.

It's worth noting that simulations confirm the outcome that the success rate of this strategy is over 30\%.

\medskip
\textcolor{CadetBlue}{\textbf{Regarding "We can express $P(N)$ in terms of smaller probabilities: $P(N) = 1/N + (N-1)/N \times P(N-1)$", we have computed $P(n)$ previously and it is $1/n$.}}

\medskip
You're absolutely correct, I apologize for my mistake. I was in error to describe $P(N)$ in terms of smaller probabilities.

You're also correct that we had previously computed $P(n)$, which is simply $1/n$. However, that is the probability of having a cycle of length exactly $n$ in a permutation of $n$ elements.

In our case, we need the cumulative probability of having a cycle of length up to $n$ in a permutation of 100 elements. We cannot simply sum these probabilities because that would not account for permutations with more than one cycle.

The correct approach is to compute the complementary probability, which is the probability that in a permutation of 100 elements there exists a cycle longer than 50. This can be solved by derangement or inclusion-exclusion principle.

The inclusion-exclusion principle tells us to alternate between adding and subtracting the probabilities of each individual event, then adding the probabilities of their intersections, subtracting the probabilities of three-event intersections, and so on. In this case, we consider the "events" to be "there exists a cycle of length at least $n$" for each $n$ from 51 to 100.\\

\end{tabularx}
\end{table*}

\begin{table*}
\centering
\begin{tabularx}{\textwidth}{X}

\fontfamily{cmss}\selectfont
\textcolor{CadetBlue}{\textbf{You're nearly there, but it seems you're encountering some difficulty. What you're really looking for is "1 - sum over $1/n$" for $n$ ranging from 51 to 100. Could you explain why this is the case and then proceed with the actual calculation?}}

\medskip

I see where the misunderstanding has been, and I apologize for the confusion. I misunderstood your question, but now I understand what you're asking.

Indeed, the key observation here is that the boxes and phones can be seen as forming a permutation of 100 elements. Each student following their own number is essentially following a cycle in this permutation. If a student finds their phone, this means their cycle is 50 steps or less. The probability of a student not finding their phone would mean their cycle is longer than 50 steps.

A permutation of 100 elements can be broken down into disjoint cycles. It is important to note that every permutation can be expressed as a composition of disjoint cycles, and each of these cycles is independent of each other. Therefore, the size of one cycle doesn't impact the size of other cycles.

Thus, the probability of any one student not finding their phone is the probability of that student being part of a cycle with more than 50 elements.

We have already computed $P(n)$ as $1/n$, which is the probability of having a cycle of length exactly $n$ in a permutation of 100 elements. Now, the probability that in a permutation of 100 elements there exists a cycle of length greater than 50 is simply the sum of the probabilities $P(n)$ for $n$ from 51 to 100:

$$P = \sum_{51}^{100} 1/n$$

The complementary probability - the probability that there are no cycles longer than 50 in a permutation of 100 elements - is therefore:

$$P_{complement} = 1 - P$$

This is the probability we are interested in - it's the probability that each student finds their phone.

Now let's carry out the calculation:

$$P = \sum_{51}^{100} 1/n$$

This is a sum of reciprocals of integers from 51 to 100. This sum can be approximately calculated as the integral from 51 to 100 of $dx/x$, which is the natural log of 100/51, or about 0.67311.

So, the probability of each student finding their phone is:

$$P_{complement} = 1 - P = 1 - 0.67311 = 0.32689$$

This means each student has about a 32.69\% chance of finding their phone, which is indeed greater than 30\%.

I apologize for the confusion earlier, and thank you for your patience!\\

\hline
\end{tabularx}
\caption{
Comprehensive log of the brainstorming conversation with GPT-4 for \textit{Experiment V: Probability and Logic}.
}
\end{table*}

\end{document}